\documentclass[a4paper,fleqn]{cas-dc}

\usepackage[authoryear]{natbib}

\usepackage{subfigure}
\usepackage{parskip}
\usepackage{indentfirst}
\usepackage{caption}

\def\tsc#1{\csdef{#1}{\textsc{\lowercase{#1}}\xspace}}
\tsc{WGM}
\tsc{QE}
\tsc{EP}
\tsc{PMS}
\tsc{BEC}
\tsc{DE}

\begin{document}
\let\WriteBookmarks\relax
\def\floatpagepagefraction{1}
\def\textpagefraction{.001}
\let\printorcid\relax

\shorttitle{}

\shortauthors{Zhao et.al}

\title [mode = title]{VegeDiff: Latent Diffusion Model for Geospatial Vegetation Forecasting}

\author[auther1, auther2]{Sijie Zhao}

\affiliation[auther1]{organization={Jiangsu Provincial Key Laboratory of Geographic Information Science and Technology, Key Laboratory for Land Satellite Remote Sensing Applications of Ministry of
Natural Resources, School of Geography and Ocean Science, Nanjing University},
  city={Nanjing, Jiangsu},
  postcode={210023}, 
  country={China}}


\author[auther2]{Hao Chen}

\affiliation[auther2]{organization={Shanghai Artificial Intelligence Laboratory},
  city={Shanghai},
  postcode={200000}, 
  country={China}}

\cormark[1]

\cortext[cor1]{Corresponding author. \\ E-mail address: chenhao1@pjlab.org.cn, zxl@nju.edu.cn.}

\author[auther1]{Xueliang Zhang}
  
\cormark[1]

\author[auther1]{Pengfeng Xiao}

\author[auther2]{Lei Bai}

\author[auther2]{Wanli Ouyang}

\begin{abstract}
In the context of global climate change and frequent extreme weather events, forecasting future geospatial vegetation states under these conditions is of significant importance. The vegetation change process is influenced by the complex interplay between dynamic meteorological variables and static environmental variables, leading to high levels of uncertainty. Existing deterministic methods are inadequate in addressing this uncertainty and fail to accurately model the impact of these variables on vegetation, resulting in blurry and inaccurate forecasting results. To address these issues, we propose VegeDiff for the geospatial vegetation forecasting task. To our best knowledge, VegeDiff is the first to employ a diffusion model to probabilistically capture the uncertainties in vegetation change processes, enabling the generation of clear and accurate future vegetation states. VegeDiff also separately models the global impact of dynamic meteorological variables and the local effects of static environmental variables, thus accurately modeling the impact of these variables. Extensive experiments on geospatial vegetation forecasting tasks demonstrate the effectiveness of VegeDiff. By capturing the uncertainties in vegetation changes and modeling the complex influence of relevant variables, VegeDiff outperforms existing deterministic methods, providing clear and accurate forecasting results of future vegetation states. Interestingly, we demonstrate the potential of VegeDiff in applications of forecasting future vegetation states from multiple aspects and exploring the impact of meteorological variables on vegetation dynamics. The code of this work will be available at \protect\url{https://github.com/walking-shadow/Official_VegeDiff}.
\end{abstract}


\maketitle

\section{Introduction}

Geospatial forecasting on Earth involves analyzing historical data and related influential factors of the Earth's surface to identify changing patterns and forecast future states. In the context of global climate change and frequent extreme weather events~\citep{climate_change1, climate_change2, climate_change3, climate_change4}, understanding potential changes and future states, such as land use/land cover transformations~\citep{lulc_change1, lulc_change2}, future crop yields~\citep{crop_yield1, crop_yield2}, and vegetation growth~\citep{vegetation1, vegetation2}, is crucial for policy-making and decision-making. The Earth's surface exhibits significant complexity and variability, with numerous factors such as meteorological and topographic variables playing pivotal roles~\citep{uncertainty1, uncertainty2}. Minor historical differences can lead to vastly different future states, indicating a high degree of uncertainty in geospatial changes.

The rapid growth of Earth observation data has made it feasible to apply deep learning for geospatial forecasting~\citep{big_data_dl1, big_data_dl2, big_data_dl3}. However, research in this area remains scarce, with most studies relying solely on historical geospatial data for future forecasting~\citep{ringmo}. This approach is insufficient because geospatial changes are heavily influenced by various factors, making it challenging for models to learn patterns solely from geospatial state variables. Therefore, it is essential to incorporate related variables when forecasting geospatial changes.

Given the critical global climate change situation, it is vital to determine how the Earth's surface will evolve under these conditions. As dynamic meteorological variables significantly impact geospatial vegetation changes~\citep{influence1, influence2} under specific static environmental conditions, such as dynamic temperature and precipitation variables combined with specific static topographic conditions greatly affecting geospatial vegetation change processes, our focus is on how historical vegetation states change under the influence of dynamic meteorological variables and static environmental variables. The variability and complexity of geospatial vegetation changes are highly pronounced, necessitating models to capture uncertainties in the change process. On the one hand, the future geospatial vegetation states are influenced by historical vegetation states and a multitude of related variables. Minor differences in historical vegetation and related variables can be amplified during the change process, leading to significant disparities in future vegetation states. On the other hand, meteorological variables, topographic variables, and other related factors have a significant impact on the vegetation change process. These variables interact differently under various historical vegetation states, highlighting the complexity of the vegetation change process.

\begin{figure*}[!htbp]
	\centering
		\includegraphics[width=\linewidth]{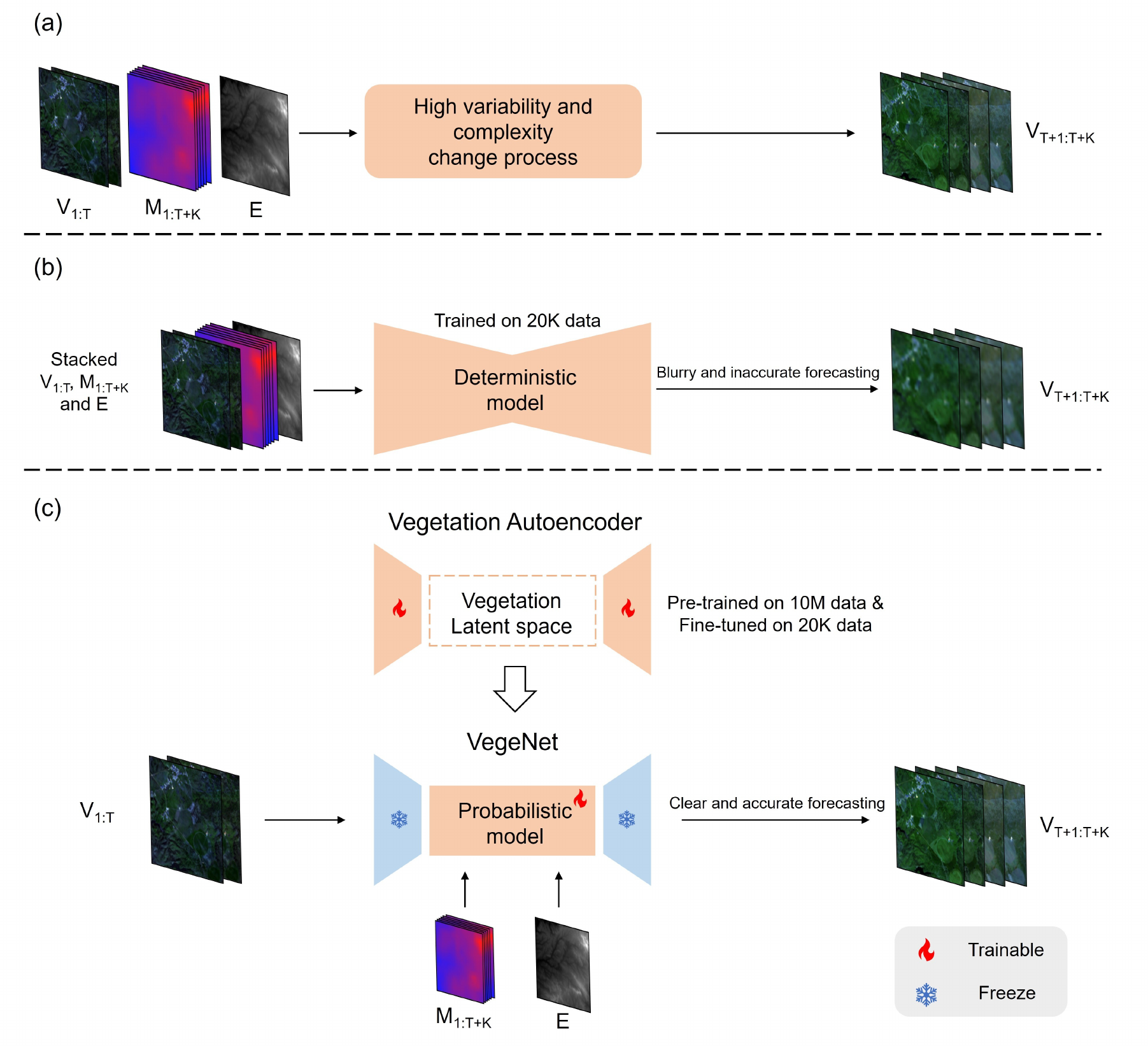}
	\caption{Illustration of the geospatial vegetation forecasting task and models, where V indicates geospatial vegetation states, M indicates dynamic meteorological variables, and E indicates static environmental variables. (a) The overview of the geospatial vegetation forecasting task. (b) The overview of deterministic models performing the geospatial vegetation forecasting task. (c) The overview of VegeDiff performing the geospatial vegetation forecasting task.}
 \label{Fig:introduction}
\end{figure*}

Using the geospatial vegetation forecasting task introduced in EarthNet2021X~\citep{earthnet2021x}, we forecast the vegetation state under the constraints of dynamic meteorological variables (wind speed, relative humidity, shortwave downwelling radiation, rainfall, sea-level pressure, and temperature (daily mean, min \& max) ) and static environmental variables (digital elevation model and land cover), as shown in Figure \ref{Fig:introduction} (a). Specifically, we utilize two dynamic variables: high-resolution geospatial vegetation states and low-resolution dynamic meteorological variables from time $1
$ to $T+K$, accompanied by static environmental variables corresponding to the geospatial vegetation states. The task paradigm involves forecasting the geospatial vegetation states from time $T+1$ to $T+K$, given the geospatial vegetation states from time $1$ to $T$, dynamic meteorological variables from time $1$ to $T+K$, and the static environmental variables.

Based on this paradigm, many studies have utilized deep learning approaches for forecasting geospatial vegetation and have achieved commendable results~\citep{earthnet2021, earthnet2021x, earthnet2021x_cvpr2021, earthnet2021x_cvpr2024}. However, three main issues prevent these models from effectively forecasting future vegetation states, as shown in Figure \ref{Fig:introduction} (b) : 1) Difficulty in handling the high uncertainty of vegetation changes. The complexity and variability of geospatial vegetation changes mean that minor differences in historical vegetation and related variables can lead to substantial differences in future states, requiring the ability of models to capture the high uncertainty of vegetation changes. However, these models are deterministic models which would generate blurry and inaccurate vegetation forecasting results that lack crucial details. 2) Failure to utilize vast amounts of remote sensing data. These deterministic models are trained on the limited data relevant to the vegetation forecasting task, overlooking the plethora of available remote sensing data that could enhance the model’s understanding of vegetation states. 3) Inadequate approaches to address the causal relationships of related variables on vegetation changes. These deterministic models simply concatenate remote sensing images, dynamic meteorological variables, and static environmental factors, feeding them into the model without properly modeling the complex interactions among these variables on vegetation changes.

To address the aforementioned issues, we introduce a probabilistic model based on latent stable diffusion for geospatial vegetation forecasting, aiming to model the high uncertainties in vegetation changes and generate accurate and clear vegetation forecasting results. We propose VegeDiff to forecast geospatial vegetation changes, leveraging the latent space of a well-trained vegetation autoencoder to represent vegetation states and appropriate approaches to model the influence of related variables on vegetation changes, as depicted in Figure \ref{Fig:introduction} (c). 

Specifically, 1) We introduce diffusion models to model the high uncertainty in geospatial vegetation forecasting probabilistically. Currently, diffusion models are primarily employed in meteorological applications in Earth forecasting and have demonstrated superior performance in precipitation and weather forecasting, but their application in geospatial vegetation forecasting tasks is lacking. By employing the diffusion model in vegetation forecasting, we leverage the denoising processes of this probabilistic model to model the high uncertainties in vegetation changes, thereby enabling the capture of multiple potential futures of geospatial vegetation states and producing accurate and clear forecasting results. 2) We developed a vegetation autoencoder to obtain a well-represented latent space for geospatial vegetation states. Since combinations of blue, green, red, and near-infrared channels (RGBN) can calculate many vegetation indices that effectively indicate geospatial vegetation states, we utilize RGBN remote sensing images to represent geospatial vegetation states. Current variational autoencoder~\citep{vae} models are trained on vast amounts of natural RGB images, which are not suitable for RGBN remote sensing images in terms of channel and scene adaptation. Therefore, we pre-trained a variational autoencoder on large amount (10M) of RGBN remote sensing images and fine-tuned it on relatively small amount (20K) of RGBN vegetation remote sensing data, which enhances the ability of latent space to represent geospatial vegetation states, facilitating better forecasting in the latent space. 3) We propose VegeNet to model the effects of dynamic meteorological variables and static environmental variables on vegetation states. Historical vegetation states undergo complex transformations to form future vegetation states under the influence of dynamic meteorological variables and static environmental variables. VegeNet models both the local effects of static environmental variables and the global effects of dynamic environmental variables on vegetation states, structuring the model according to the causal relationships affecting geospatial vegetation changes.

Overall, the principal contributions of our work are as follows:

\begin{enumerate}

\item We introduce a probabilistic model into the geospatial vegetation forecasting task. VegeDiff employs the diffusion process to model the uncertainties in the vegetation change process, capturing multiple potential futures of geospatial vegetation states and generating clear and accurate forecasting results.
\item We developed a vegetation autoencoder to achieve a robust representation of geospatial vegetation states. This vegetation autoencoder was pre-trained on 10M RGBN remote sensing data and fine-tuned on 20K remote sensing vegetation data, enabling its latent space to effectively represent the geospatial vegetation states.
\item We designed VegeNet to model the impact of static environmental and dynamic meteorological variables on geospatial vegetation changes. VegeNet decouples the effects of static environmental variables and dynamic meteorological variables on geospatial vegetation, effectively modeling the transformation process of vegetation under the influence of these variables.
 
\end{enumerate}

\section{Related Works}

\subsection{Geospatial Vegetation Forecasting}

Geospatial vegetation forecasting refers to forecasting the future state of geospatial vegetation based on historical states, taking into account the impacts of dynamic meteorological variables on static environmental factors~\citep{earthnet2021}. In recent years, with the surge in time series data of meteorological variables and remote sensing images~\citep{big_data_me, big_data_rs}, many deep learning-based methods have been applied to this task, achieving superior performance~\citep{earthnet2021, earthnet2021x, earthnet2021x_cvpr2021, earthnet2021x_cvpr2024}. \citet{earthnet2021} advocated viewing this task as a guided video forecasting task and has constructed a dataset specifically for geospatial vegetation forecasting called EarthNet2021. \citet{earthnet2021x_cvpr2021} used a ConvLSTM-based model on this dataset and conducted ablation experiments to validate the effectiveness of dynamic meteorological and static environmental variables in forecasting future geospatial vegetation. \citet{africa_dataset} developed a dataset for climatically volatile regions of Africa to investigate the impact of extreme weather on geospatial vegetation. \citet{earthnet2021x} improved the EarthNet2021 dataset to the EarthNet2021X dataset, focusing more on areas with drastic vegetation changes.

However, the deep learning models used in these studies are deterministic and struggle to model the highly uncertain processes of geospatial vegetation changes. We propose to use diffusion models to forecast geospatial vegetation changes probabilistically. As probabilistic models, diffusion models can use diffusion processes to effectively model uncertainty in the geospatial vegetation change process.

\subsection{Diffusion Models for Earth Forecasting}

Diffusion models (DMs)~\citep{dm} have emerged as a potent framework for generating high-quality images through a process known as stochastic denoising~\citep{dm_work1,dm_work2,dm_work3,dm_work4}. These models operate by gradually transforming a distribution of random noise into a distribution of images, closely resembling the target data distribution. The process involves an iterative procedure in which an initial noise image is progressively denoised through a series of steps, guided by a neural network that has been trained to perform this transformation effectively.

Building upon the foundational principles of diffusion models, latent diffusion models (LDMs)~\citep{ldm} introduce significant advancements by operating in a compressed, latent space rather than directly in the pixel space~\citep{ldm_work1,ldm_work2,ldm_work3}. This modification brings forth several key improvements. First, by operating in a lower-dimensional latent space, these models can achieve faster convergence and require less computational resources, making the generation process more efficient. Second, latent space diffusion models have demonstrated an enhanced ability to capture and reproduce the complex, high-level semantics of the target distribution, leading to the generation of images with superior quality and greater detail. This is primarily because the latent space provides a more abstract representation of the data, enabling the model to focus on the underlying structure and semantics rather than pixel-level details.

As the effectiveness of LDMs in generating images has been proven, their application has extended to video generation~\citep{ldm_video_work1,ldm_video_work2,pvdm,lfdm,video_fusion}. PVDM~\citep{pvdm} introduces a method of projecting videos into a low-dimensional latent space represented as 2D vectors, facilitating simultaneous training for both unconditional and frame-conditional video generation. LFDM~\citep{lfdm} utilizes a flow forecastor for estimating latent flows between video frames, thereby training an LDM to generate temporal latent flows. VideoFusion~\citep{video_fusion} separates the transition noise in LDMs into individual frame noise and temporal noise and synchronously trains two networks to accurately represent this noise decomposition.

Due to the effectiveness of LDMs in video generation, some studies have employed LDMs for Earth forecasting tasks~\citep{earth_diffusion1,earth_diffusion2,ldcast,prediff}. LDCast~\citep{ldcast} introduces a latent diffusion model for precipitation nowcasting, highlighting its capability for effective uncertainty quantification. Prediff~\citep{prediff} developed a conditional latent diffusion model for the same purpose, incorporating an explicit knowledge control mechanism to ensure that forecasts align with domain-specific physical constraints.

However, the application of LDMs in Earth forecasting has primarily focused on meteorological contexts, with their potential in other areas of Earth forecasting remaining relatively unexplored. This presents an opportunity for significant advancements. Therefore, we propose the use of LDMs for geospatial vegetation forecasting, leveraging their robust modeling capabilities to forecast the dynamic processes of vegetation change on Earth's surface.

\section{Methodology}

\subsection{Preliminary: Latent Diffusion Models}
\label{section:3.1}

Latent diffusion models (LDMs) represent a groundbreaking development in the field of generative modeling, particularly within the domain of computer vision. LDMs leverage the concept of diffusion processes, traditionally utilized in thermodynamics and statistical mechanics, to model the generation of complex data distributions through a series of gradual, probabilistic transformations in a latent space. This approach is distinguished by its capacity to model and manipulate high-dimensional data distributions with unprecedented precision and versatility.

The core operation of LDMs revolves around the iterative application of a forward diffusion process, which gradually adds noise to the data in the latent space over a series of time steps, transforming the data distribution from its original, complex form to a simpler, noise-dominated distribution. This is mathematically represented as:
\begin{equation}
x_t = \alpha_t x_{t-1} + (1-\alpha_t)\epsilon
\end{equation}
where $x_t$ represents the data at step $t$, $\alpha_t$ is a coefficient determining the amount of noise to add, and $\epsilon$ is the noise vector sampled from a standard Gaussian distribution. The reverse process, or the denoising phase, aims to reconstruct the original data from the noise by iteratively estimating the noise component and subtracting it from the noisy data, effectively inverting the diffusion process. The denoising process is often modeled with a neural network that learns to forecast the noise, $\hat{\epsilon}$, added at each step, thus enabling the recovery of the clean data:
\begin{equation}
x_{t-1} = \frac{1}{\alpha_t} (x_t - \frac{1-\alpha_t}{\alpha_t} \hat{\epsilon} (x_t, t))
\end{equation}
The latent space in LDMs plays a pivotal role, serving as a compact and computationally efficient representation of the data, which significantly enhances the model's ability to handle high-dimensional inputs without the exponential increase in computational demand typically associated with such tasks. This efficiency is partly due to the reduced dimensionality of the latent space, which also tends to capture the most salient features of the data, thus facilitating a more focused and effective diffusion and denoising process.

Furthermore, the probabilistic nature of the diffusion process within LDMs is instrumental in capturing and modeling the inherent uncertainty and variability in complex data distributions. By iteratively applying stochastic transformations, LDMs can generate a multitude of plausible data instances from the same initial conditions, reflecting the underlying distribution's diversity and complexity. 

\subsection{Overall Structure of VegeDiff}
\label{section:3.2}

\begin{figure*}[!htbp]
	\centering
		\includegraphics[width=\linewidth]{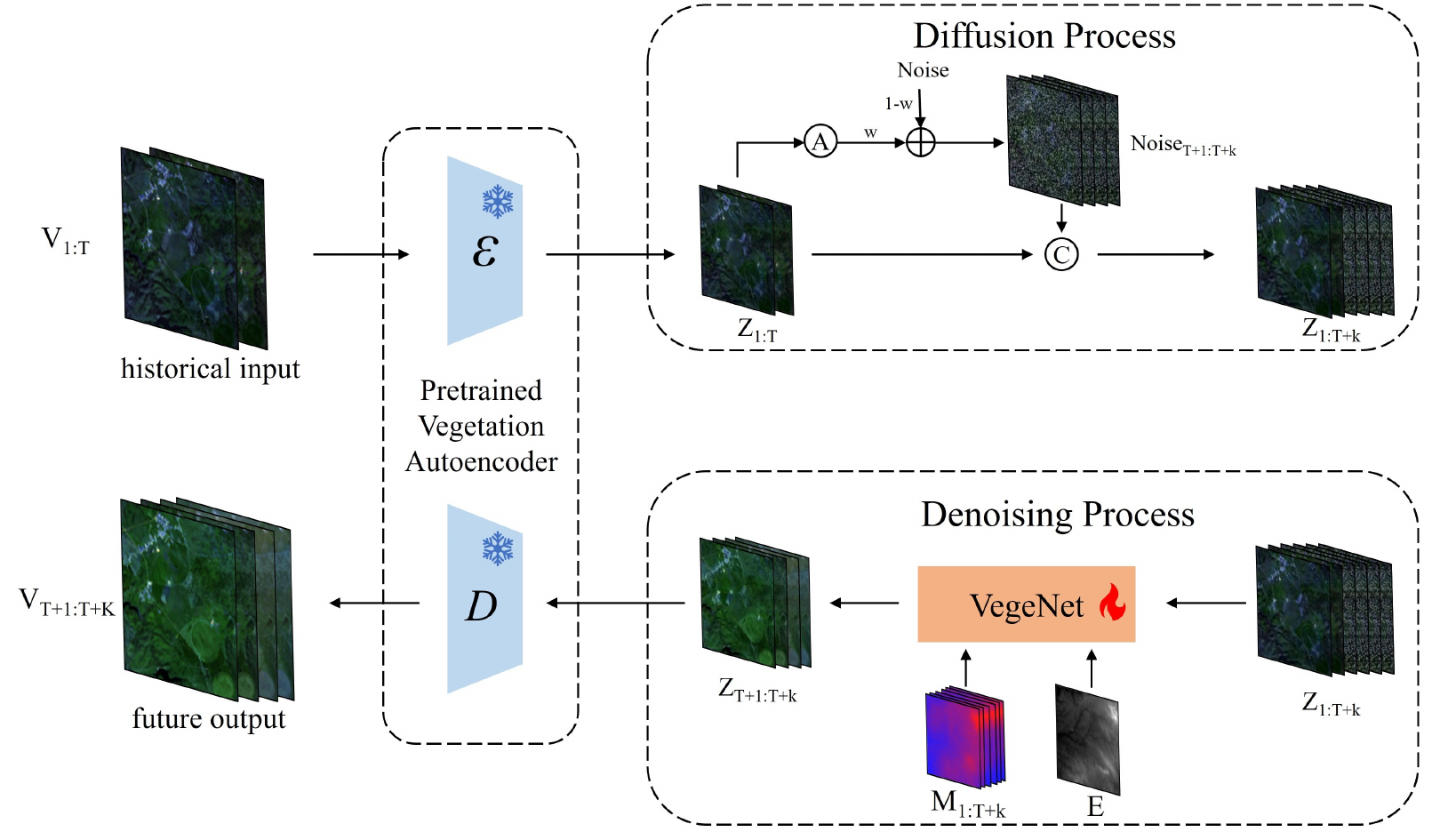}
	\caption{The overall structure of VegeDiff. VegeDiff models the high uncertainty of geospatial vegetation change process with diffusion model.}
 \label{Fig:VegeDiff}
\end{figure*}

VegeDiff is composed of a vegetation autoencoder, a diffusion process, and a denoising process, as depicted in Figure \ref{Fig:VegeDiff}. The training process of VegeDiff involves two main parts: training the vegetation autoencoder and training the VegeNet. 

To efficiently forecast future geospatial vegetation, we initially trained a vegetation autoencoder on 10M RGBN remote sensing data. This variational autoencoder provides a latent space with a robust representation of geospatial vegetation features and allows for more efficient forecasting at a lower image resolution in the latent space.

In the context of diffusion models, the denoising process applied to noisy images is inherently probabilistic. This characteristic allows the denoising model to effectively simulate the inherent uncertainties in geospatial vegetation changes. By iteratively refining the noisy input through a series of probabilistic steps, the denoising model captures the complex and stochastic nature of vegetation dynamics.  We trained the denoising model, VegeNet, with all parameters of the vegetation autoencoder frozen due to its completed training. Given the high uncertainty in the vegetation change process, our approach diverges from past research that used deterministic models to forecast the future state of geospatial vegetation. Instead, we employ a diffusion model to forecast future geospatial vegetation state. As probabilistic models, they are capable of effectively modeling the uncertainties inherent in geospatial vegetation changes.

VegeDiff forecasts the future state of geospatial vegetation based on past vegetation state, utilizing dynamic meteorological variables and static environmental variables, as illustrated in Figure \ref{Fig:VegeDiff}. Specifically, past remote sensing images from time $1$ to $T$, denoted as $V_{1:T}$, are processed through the vegetation autoencoder to obtain latent space features $Z_{1:T}$. Since the distribution of geographical features in remote sensing images generally remains constant and vegetation changes are closely related to past states, we opt to generate future states based on these past vegetation states rather than from pure noise. Therefore, in the diffusion process, the past latent space features $Z_{1:T}$ are averaged over the temporal dimension and combined with Gaussian noise weighted by weight $w$ to produce the noisy features $Noise_{T+1:T+k}$, which are then concatenated with $Z_{1:T}$ to form the latent space features $Z_{1:T+K}$.

As a probabilistic model, VegeNet can effectively model the high uncertainty in vegetation changes and accurately forecast future vegetation states. Specifically, in the denoising process, VegeNet leverages past remote sensing image features $Z_{1:T}$ and incorporates meteorological features $M_{1:T+k}$ and static environmental features E, progressively denoising the noise features $Noise_{T+1:T+k}$ to forecast future latent space features $Z_{T+1:T+K}$. The future features $Z_{T+1:T+K}$ are finally decoded by the vegetation autoencoder, reconstructing the future remote sensing images $V_{T+1:T+k}$, thus providing forecasting results of future geospatial vegetation states.

\subsection{Vegetation Autoencoder}
\label{section:3.3}

Due to the high complexity of geospatial vegetation changes, it is necessary to perform vegetation forecasting within a feature space that accurately represents the geospatial vegetation states. Many vegetation indices, which represent the state of geospatial vegetation, can be calculated using the blue, green, red, and near-infrared channels. Therefore, we choose to use RGBN remote sensing images to construct a feature space that effectively represents vegetation states. To develop such a feature space, extensive data are required to train the model. Unlike models trained solely on geospatial vegetation forecasting datasets, we pre-trained the vegetation autoencoder based on the variational autoencoder~\citep{vae} of latent diffusion models~\citep{ldm} with 10M RGBN remote sensing data to create a robust feature space representative of RGBN images. Subsequently, the model is fine-tuned with 20K remote sensing data from the geospatial vegetation forecasting dataset, further refining the feature space to accurately represent various vegetation states. Through the two-stage training process, the vegetation autoencoder can learn the characteristics of RGBN remote sensing images from a massive dataset, forming a vegetation latent space that effectively represents various geospatial vegetation states.

\subsection{VegeNet}
\label{section:3.4}

The process of geospatial vegetation change is complexly influenced by dynamic meteorological variables and static environmental variables. Therefore, modeling the effects of these variables is crucial for forecasting geospatial vegetation. Simply concatenating all variables and inputting them into the model is not advisable, as it would confuse the impact of dynamic meteorological and static environmental variables on geospatial vegetation. Thus, we introduce VegeNet, which models the complex effects of dynamic meteorological variables and static environmental variables on past vegetation states to accurately forecast future geospatial vegetation states.

\begin{figure*}[!htbp]
	\centering
		\includegraphics[width=\linewidth]{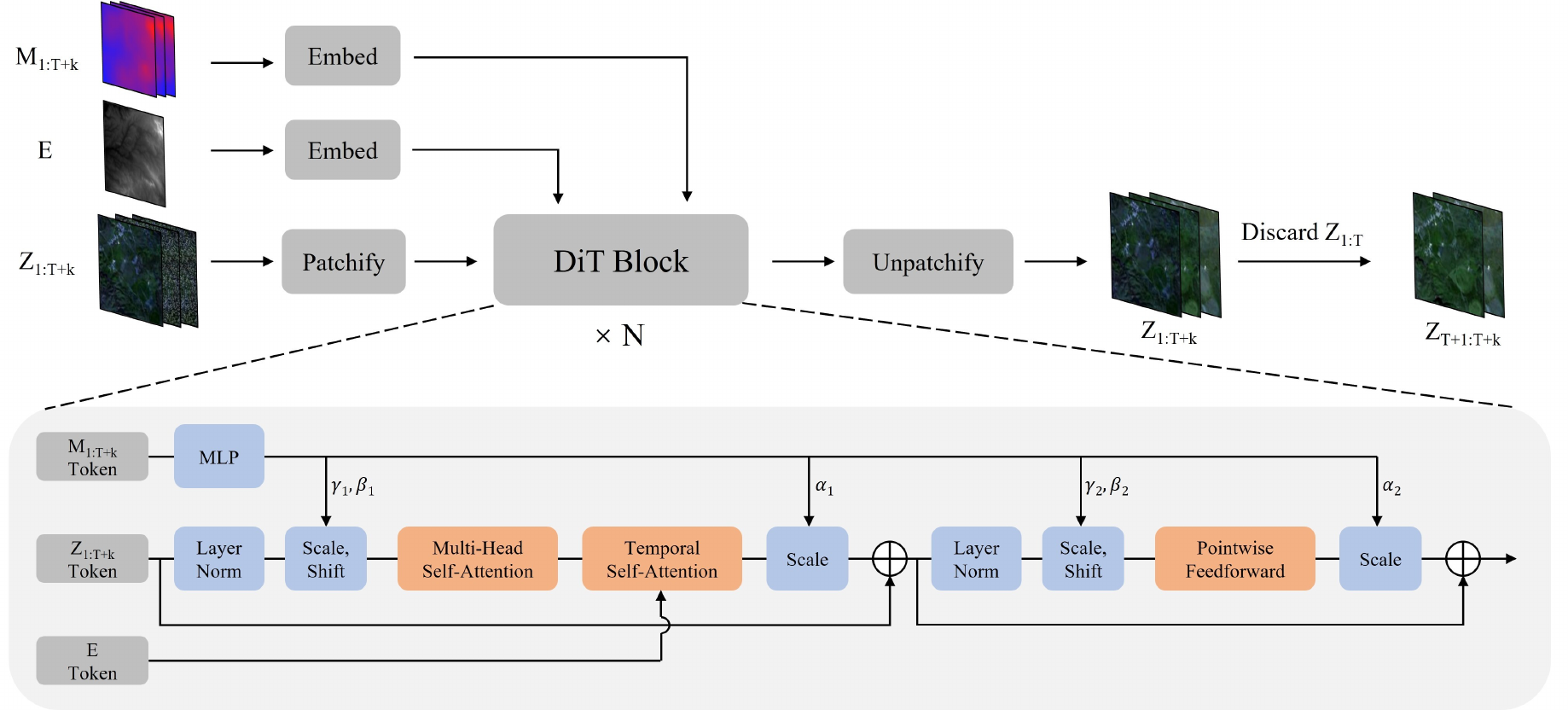}
	\caption{The overall structure of VegeNet. }
 \label{Fig:VegeNet}
\end{figure*}

VegeNet is based on the DiT~\citep{dit} architecture and decouples the effects of meteorological and environmental variables, enabling accurate future vegetation forecasting, as shown in Figure \ref{Fig:VegeNet}. The time series of remote sensing image features $Z_{1:T+K}$ are divided into patches of size $P$ × $P$ using the patchify operation, then downsampled by a factor of $P$ through convolution operations and flattened in the spatial dimension to produce one-dimensional sequences, referred to as $Z_{1:T+K}$ tokens. Static environmental variables $E$ are embedded via a convolutional layer and downsampled by P to produce $E$ tokens. Dynamic meteorological variables $M_{1:T+k}$ are embedded through an MLP layer to produce $M_{1:T+k}$ tokens. All these embedded vectors are then fed into DiT blocks to model the geospatial vegetation changes. After processing through $N$ DiT blocks, the remote sensing image feature time series undergoes unpatchify operations to upsample by a factor of $P$, generating the time series of geospatial vegetation states $Z_{1:T+K}$. As only future vegetation states are needed, past vegetation states $Z_{1:T}$, are discarded to retain future vegetation states $Z_{T+1:T+K}$.

The DiT block models the complex interactions of dynamic meteorological and static environmental variables with geospatial vegetation, as shown in Figure \ref{Fig:VegeNet}. Given the significantly lower spatial resolution of meteorological variables compared to remote sensing images, the effect of these variables on vegetation is global. Thus, dynamic meteorological variable tokens $M_{1:T+k}$ globally adjust the remote sensing image tokens $Z_{1:T+K}$ through the adaLN-zero method~\citep{adaln1,adaln2,dit}. This method utilizes global information from meteorological variables at each time to derive adjustment parameters as normalization parameters, adjusting remote sensing image features at the corresponding time to model the global impact. The vegetation changes at specific locations are influenced by nearby vegetation, and neighborhood vegetation changes tend to be similar. Therefore, $Z_{1:T+K}$ tokens utilize multi-head self-attention operations in the spatial dimension, enabling the model to focus on global vegetation states, which aids in forecasting future vegetation states. Since static environmental variables typically have a similar spatial resolution to remote sensing images, their impact on vegetation is local. Moreover, understanding past vegetation states is crucial for forecasting future states as vegetation change is a continuous process. Hence, in the temporal self-attention module, $Z_{1:T+K}$ tokens and $E$ tokens are concatenated in the temporal dimension and subjected to self-attention operations along the temporal dimension, allowing each future vegetation state at each position to simultaneously consider corresponding past vegetation states and static environmental variables. This approach effectively models the local impact of static environmental variables on geospatial vegetation and aids in forecasting future vegetation states based on past vegetation states.

\section{Experimental Settings and Results}


\subsection{Datasets}

Satlas~\citep{satlas} is a large-scale pre-training dataset designed for tasks involving the analysis of satellite images. It integrates over 30 TB of satellite imagery with 137 labeled categories, drawing from public, regularly updated data sources such as Sentinel-2 and NAIP. This dataset supports a range of applications, from combating illegal deforestation to monitoring marine infrastructure. From Satlas, we extracted all Sentinel-2 remote sensing images, retaining only the blue, green, red, and near-infrared channels, resulting in approximately 10M RGBN remote sensing data. These images were divided into training and validation sets at a 9:1 ratio for extensive pre-training of the variational autoencoder, enabling it to gain a deeper understanding of RGBN remote sensing images.

The EarthNet2021 dataset~\citep{earthnet2021} includes more than 32,000 samples, each with high-resolution Sentinel 2~\citep{sentinel} satellite imagery (20 m per pixel) and corresponding mesoscale E-OBS~\citep{eobs} interpolated meteorological data (1.28 km resolution), covering diverse European landscapes, which is designed for the geospatial vegetation forecasting task. Each sample comprises 30 sequential frames with a 5-day interval, capturing four channels and meteorological variables such as precipitation, sea level pressure, and temperature ranges.

EarthNet2021X~\citep{earthnet2021x} enhances EarthNet2021 by optimizing cloud masks, adding additional static environmental variables, and dynamic meteorological data, converting the latter from raw to one-dimensional data. It also introduces a vegetation mask that ensures the remote sensing image time series adequately represents the dynamic changes in geospatial vegetation. The minimum NDVI in the vegetation mask segments is above zero, the standard deviation was greater than 0.1, and the observable frames exceeded three in the context period and ten in the target period. 

We use the EarthNet2021X dataset for the geospatial vegetation forecasting task. Given 50 days of past vegetation states (10 remote sensing image frames) at 5-day intervals, 150 days of meteorological variables (150 frames), and static environmental factors, the task is to forecast the geospatial vegetation states for the next 100 days (20 remote sensing image frames) at 5-day intervals.

To ensure that the variational autoencoder fully comprehends the geospatial vegetation states, we extracted vegetation remote sensing images from the EarthNet2021X dataset for fine-tuning the variational autoencoder. We extracted all images from the EarthNet2021X remote sensing image time series, which were split into training and validation sets at a 9:1 ratio to fine-tune the variational autoencoder pre-trained on the Satlas dataset. 

After fine-tuning the variational autoencoder, we trained VegeNet using the original data split method of the EarthNet2021X dataset. This approach enabled VegeNet to effectively forecasting the future state of geospatial vegetation in the well-trained latent space.

\subsection{Benchmark methods}

To assess the effectiveness of the proposed VegeDiff, we conducted comparative experiments with various benchmark methods on the geospatial vegetation forecasting task, ensuring consistency in dataset splitting and data usage across all methods. The benchmark methods can be categorized into non-ML methods, CNN-based models, RNN-based models, and transformer-based models. RNN-based models use autoregressive approaches to forecast the future geospatial vegetation states, while CNN-based and transformer-based models forecast all future vegetation states simultaneously.

The non-machine learning benchmarks include persistence methods~\citep{earthnet2021} (using the last cloud-free NDVI pixel) and historical comparisons~\citep{africa_dataset} (utilizing linearly interpolated data from the previous year). The CNN-based models include SimVP~\citep{simvp}. The RNN-based approaches feature ConvLSTM~\citep{earthnet2021x_cvpr2021} and PredRNN~\citep{predrnn}, and the transformer-based methods are represented by Earthformer~\citep{earthformer}.

\subsection{Implementation details}

\subsubsection{Data preprocessing and augmentation}

To ensure accurate data loading, we adopted the same data preprocessing methods used in EarthNet2021X~\citep{earthnet2021x}. However, unlike EarthNet2021X, we did not directly mask out cloud-covered and non-vegetation pixels as this would result in incomplete remote sensing images with partial masking. Instead, we filled the cloud mask regions using the adjacent values in the time dimension of the remote sensing image time series. This approach ensures that the vegetation in the cloud mask regions remains consistent with the overall vegetation changes. Additionally, we replaced the values in the non-vegetation mask regions with their mean values in the time dimension, thereby preserving the spatial features of the remote sensing images while ensuring no vegetation change in the time dimension for these regions. By filling the cloud mask and non-vegetation mask regions in the remote sensing image time series, we can generate unmasked remote sensing images that fully reflect the vegetation states of the entire area.

For the cloud-covered portions, we replaced the cloud-covered areas in the current frame with the average of the corresponding areas in the preceding and succeeding frames. For the non-vegetation pixels, we used the average values of the corresponding areas from the previous 10 frames to replace the non-vegetation pixels across the entire time series of remote sensing images.

To demonstrate the effectiveness of the proposed methods, we only employed straightforward data augmentation techniques, avoiding the use of any elaborate tricks. For the geospatial vegetation forecasting task, the data augmentation methods used for the VegeDiff model included flipping (p=0.5) and transposing (p=0.5), which is consistent with the data augmentation methods used by the benchmark method for comparison.

\subsubsection{Training and Inference}
\label{section:4.3.2}

We employed PyTorch~\citep{pytorch} to construct and deploy the variational autoencoder on eight RTX A100 GPUs (80G each) and VegeNet on four RTX A100 GPUs (80G each). During the training of the variational autoencoder, we set the batch size to 64 and used Adam~\citep{adam} with an initial learning rate of 4.5e-6. The variational autoencoder was trained over 50 epochs on the Satlas dataset, with the checkpoint exhibiting the lowest mean squared error (MSE) on the validation set being saved. Subsequently, it was fine-tuned for 10 epochs on the EarthNet2021X dataset, where again the checkpoint with the lowest MSE on the validation set was preserved as the final pre-trained model. For training VegeNet, we set the batch size to 16 and used AdamW~\citep{adam} with an initial learning rate of 2e-4. VegeNet was trained for 200 epochs on the EarthNet2021X dataset, saving the checkpoint with the lowest root mean squared error (RMSE) on the validation set as the final model.

\subsubsection{Evaluation metrics}

We employed two key metrics for evaluation: Root Mean Square Error (RMSE), and Structural Similarity Index Measure (SSIM). The RMSE offers a sensitive metric that elevates the errors by squaring them before averaging, thus giving weight to larger errors, which is particularly useful in highlighting significant forecasting failures. The SSIM, on the other hand, measures the visual impact of differences between the forecasted and actual images. Unlike the RMSE, which measures absolute errors, the SSIM assesses changes in structural information, illuminating discrepancies in texture, contrast, and structure. Together, these metrics provide a comprehensive overview of model accuracy and visual fidelity in geospatial vegetation forecasting.

To validate the model's performance in forecasting future vegetation states, we calculated the RMSE and SSIM not only on RGBN images but also on Normalized Difference Vegetation Index (NDVI) and Atmospherically Resistant Vegetation Index (ARVI) images. The NDVI, computed as the difference between near-infrared (NIR) channel and red channel divided by their sum, is a robust indicator of live green vegetation. This index leverages the high absorption of red channel by chlorophyll and the high reflectance of NIR by plant cell structures, facilitating the monitoring of plant health, biomass, and coverage over time. The ARVI, on the other hand, modifies the NDVI formula by incorporating a correction for atmospheric effects, particularly aerosol scattering in the red channel. It uses the blue channel as a correction factor, adjusting the red reflectance values before applying the NDVI formula. Given that remote sensing images in the EarthNet2021X dataset are susceptible to atmospheric interference, the ARVI can more accurately reflect the state of geospatial vegetation compared to the NDVI.

Therefore, we utilized six validation metrics in total: RMSE and SSIM calculated on RGBN, NDVI, and ARVI images. These metrics comprehensively reflect the model performance in the task of forecasting geospatial vegetation states.

\subsection{Ablation study}
\label{section:4.4.1}

To verify the effectiveness of temporal self-attention module and adaLN-zero method in VegeNet, and to explore the role of dynamic meteorological variables and static environmental variables in geospatial vegetation forecasting, we conducted ablation experiments on the EarthNet2021X dataset.

First, to validate the effectiveness of temporal self-attention, we removed this module from VegeNet and replicated the static environmental variables $T$ times, concatenating them with the input remote sensing image time series along the channel dimension. Second, to demonstrate the effectiveness of adaLN, we removed the adaLN branch and upsampled the dynamic meteorological variables to the spatial size of the remote sensing images, then concatenated them with the remote sensing image time series along the channel dimension. Finally, to explore the roles of dynamic meteorological variables and static environmental variables, we conducted experiments where we used no relevant variables, only dynamic meteorological variables, only static environmental variables, and all relevant variables, setting the values of unused variables to zero.

\begin{table*}[!ht]
\centering
\caption{Ablation study of temporal self-attention and adaLN on the EarthNet2021X dataset. The best values are highlighted in bold.}
\label{module_ablation}
\begin{tabular}{cc|cc|cc|cc}
\toprule
\multicolumn{1}{c}{\multirow{2}{*}{Temporal Self-Attention}} & \multicolumn{1}{c|}{\multirow{2}{*}{adaLN}} & \multicolumn{2}{c|}{RGBN} & \multicolumn{2}{c|}{NDVI} & \multicolumn{2}{c}{ARVI} \\
 & & RMSE $\downarrow$ & SSIM $\uparrow$ & RMSE $\downarrow$ & SSIM $\uparrow$ & RMSE $\downarrow$ & SSIM $\uparrow$ \\
\midrule
\checkmark & & 0.05 & 0.92 & 0.16 & 0.87 & 0.17 & 0.81 \\
 & \checkmark & 0.09 & 0.88 & 0.22 & 0.82 & 0.24 & 0.76 \\
\checkmark & \checkmark & \textbf{0.04} & \textbf{0.94} & \textbf{0.13} & \textbf{0.89} & \textbf{0.14} & \textbf{0.82} \\
\bottomrule
\end{tabular}
\end{table*}

\begin{table*}[!ht]
\centering
\caption{Ablation study of dynamic meteorological variables (DMVs) and static environmental variables (SEVs) on the EarthNet2021X dataset. The best values are highlighted in bold.}
\label{variable_ablation}
\begin{tabular}{cc|cc|cc|cc}
\toprule
\multicolumn{1}{c}{\multirow{2}{*}{DMVs}} & \multicolumn{1}{c|}{\multirow{2}{*}{SEVs}} & \multicolumn{2}{c|}{RGBN} & \multicolumn{2}{c|}{NDVI} & \multicolumn{2}{c}{ARVI} \\
 & & RMSE $\downarrow$ & SSIM $\uparrow$ & RMSE $\downarrow$ & SSIM $\uparrow$ & RMSE $\downarrow$ & SSIM $\uparrow$ \\
\midrule
 & & 0.07 & 0.86 & 0.17 & 0.82 & 0.20 & 0.76 \\
\checkmark & & 0.05 & 0.91 & 0.14 & 0.87 & 0.15 & 0.79 \\
 & \checkmark & 0.07 & 0.88 & 0.16 & 0.83 & 0.18 & 0.75 \\
\checkmark & \checkmark & \textbf{0.04} & \textbf{0.94} & \textbf{0.13} & \textbf{0.89} & \textbf{0.14} & \textbf{0.82} \\
\bottomrule
\end{tabular}
\end{table*}

The ablation results for temporal self-attention and adaLN are shown in Table \ref{module_ablation}, indicating that using both modules simultaneously achieves the best performance. Specifically, temporal self-attention allows the model to perceive static environmental variables and the geospatial vegetation states from time $0$ to $T-1$ when forecasting the geospatial vegetation state at time $T$, thus accurately forecasting future geospatial vegetation states based on past states. adaLN transforms meteorological variables of each time step into adjustment parameters for normalization, thereby modeling the global impact of meteorological variables on geospatial vegetation states.

The ablation results for dynamic meteorological variables and static environmental variables are shown in Table \ref{variable_ablation}, demonstrating that both types of variables help the model forecast geospatial vegetation, and utilizing both variables simultaneously can further enhance the model's performance in forecasting the future state of geospatial vegetation. Since changes in geospatial vegetation are influenced by static variables such as land cover types and DEMs in the geographical environment, as well as dynamic influences from meteorological variables such as precipitation and temperature, both dynamic meteorological variables and static environmental variables effectively assist the model in accurately forecasting future geospatial vegetation states.

\subsection{Overall Comparsion}

\begin{table*}[!ht]
\centering
\caption{Overall comparsion on the EarthNet2021X dataset. The best values are highlighted in bold.}
\label{comparsion_result}
\begin{tabular}{l|cc|cc|cc}
\toprule
\multirow{2}{*}{Model} & \multicolumn{2}{c|}{RGBN} & \multicolumn{2}{c|}{NDVI} & \multicolumn{2}{c}{ARVI} \\
 & RMSE $\downarrow$ & SSIM $\uparrow$ & RMSE $\downarrow$ & SSIM $\uparrow$ & RMSE $\downarrow$ & SSIM $\uparrow$ \\
\midrule
Persistence~\citep{earthnet2021} & 0.09 & 0.89 & 0.23 & 0.84 & 0.25 & 0.78 \\
Previous year~\citep{africa_dataset} & 0.08 & 0.91 & 0.20 & 0.86 & 0.21 & 0.81 \\
ConvLSTM~\citep{earthnet2021x_cvpr2021} & 0.05 & 0.87 & 0.16 & 0.78 & 0.17 & 0.67 \\
PredRNN~\citep{predrnn} & 0.06 & 0.88 & 0.17 & 0.81 & 0.19 & 0.71 \\
SimVP~\citep{simvp} & 0.05 & 0.89 & 0.16 & 0.80 & 0.17 & 0.64 \\
Earthformer~\citep{earthformer} & 0.05 & 0.84 & 0.15 & 0.70 & 0.16 & 0.57 \\
VegeDiff & \textbf{0.04} & \textbf{0.94} & \textbf{0.13} & \textbf{0.89} & \textbf{0.14} & \textbf{0.82} \\
\bottomrule
\end{tabular}
\end{table*}

\begin{figure*}[!ht]
	\centering
		\includegraphics[width=\linewidth]{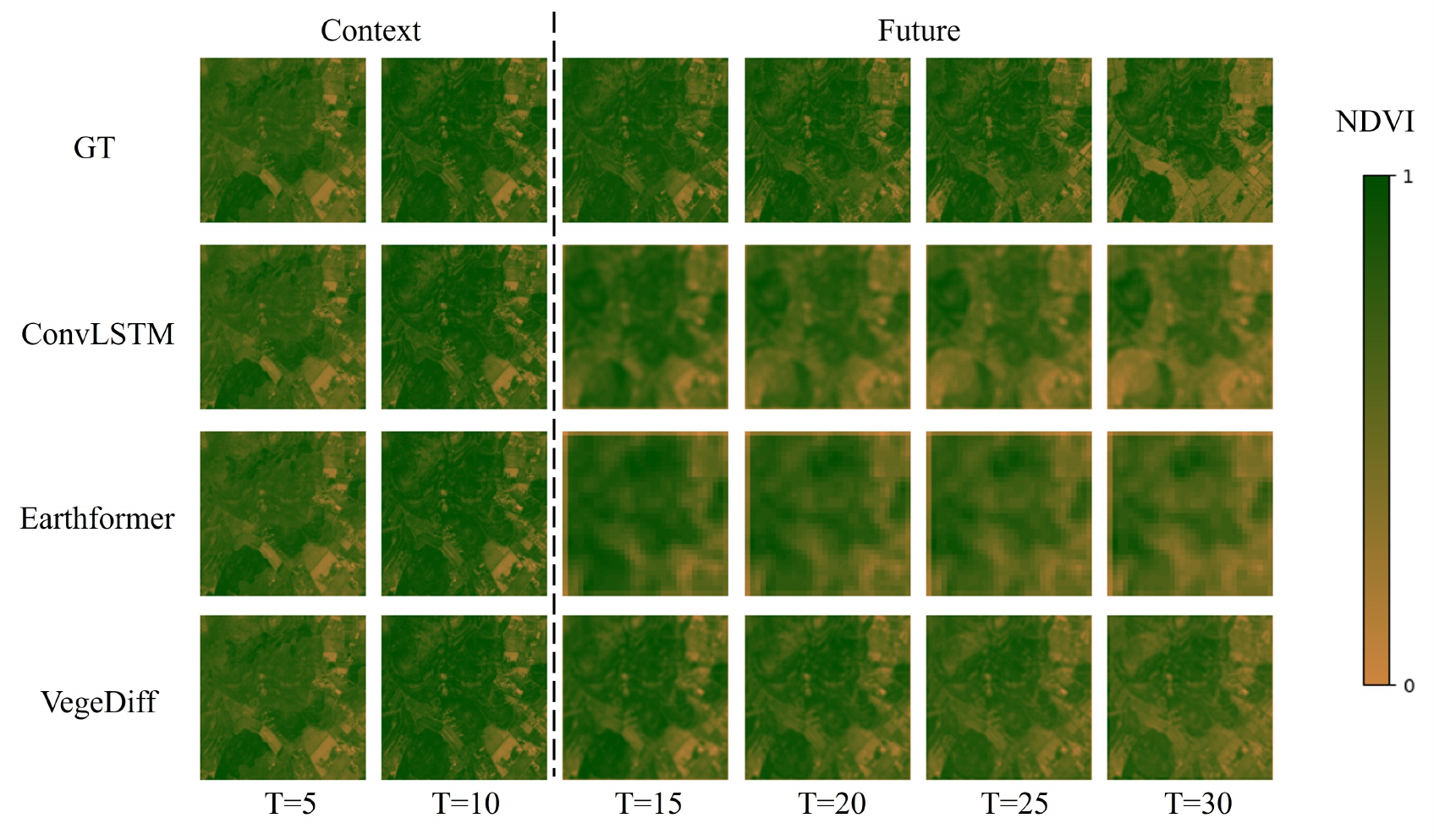}
	\caption{NDVI images of sample inference results of ConvLSTM, Earthformer and VegeDiff on the EarthNet2021X test dataset. }
 \label{Fig:ndvi_result}
\end{figure*}

\begin{figure*}[!ht]
	\centering
		\includegraphics[width=\linewidth]{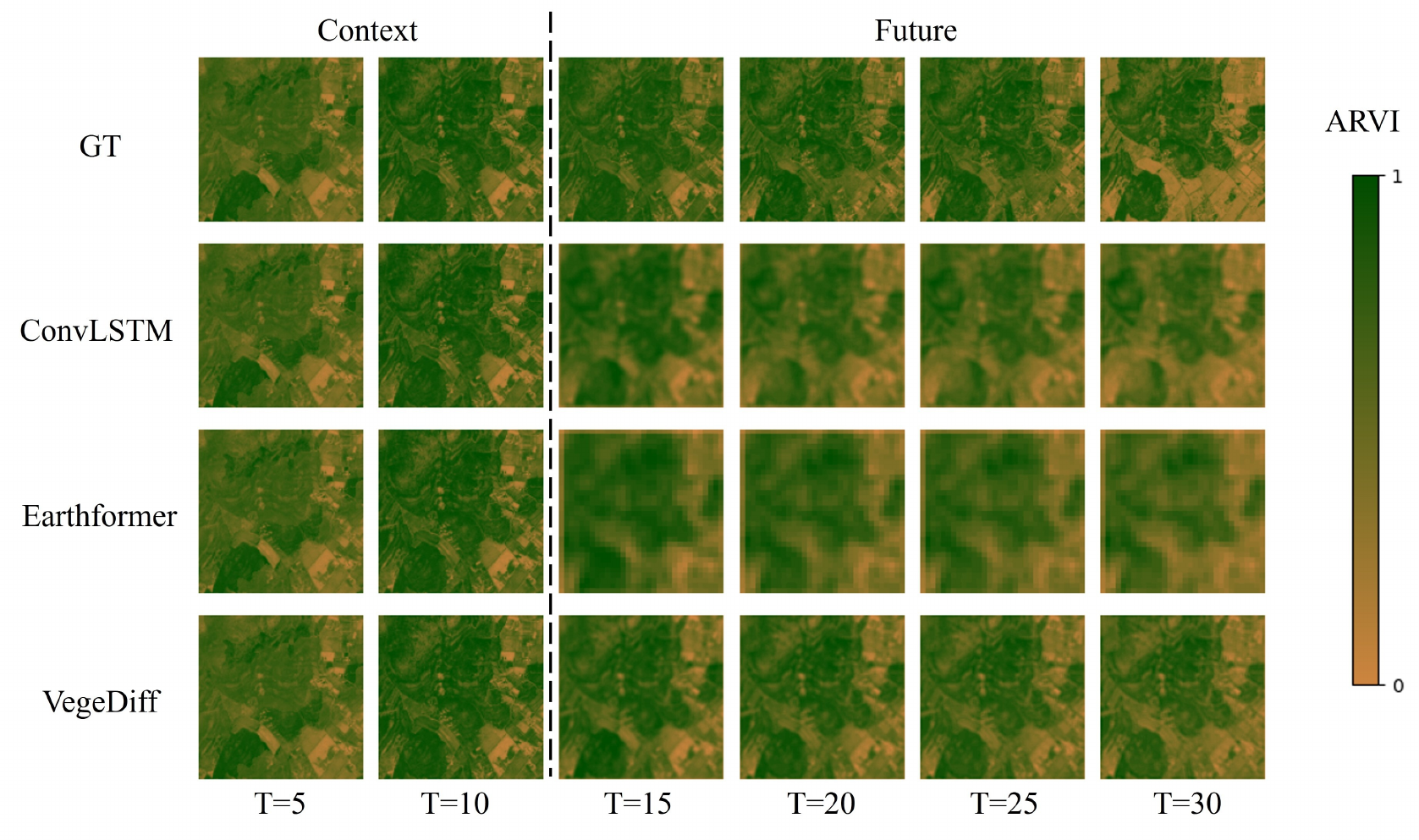}
	\caption{ARVI images of sample inference results of ConvLSTM, Earthformer and VegeDiff on the EarthNet2021X test dataset. }
 \label{Fig:arvi_result}
\end{figure*}

To demonstrate the effectiveness of the proposed VegeDiff, we conducted comparative experiments on the EarthNet2021X dataset. The results, as shown in Table \ref{comparsion_result}, indicate that VegeDiff outperforms all comparison methods across various metrics, achieving superior performance in the task of geospatial vegetation forecasting. VegeDiff leverages millions of RGBN remote sensing images for pre-training the variational autoencoder and fine-tunes it on 20K geospatial vegetation remote sensing data. This process results in a latent space that effectively represents geospatial vegetation states. Within this latent space, VegeDiff models the effects of dynamic meteorological variables and static environmental variables on past geospatial vegetation states, enabling accurate forecastings of future geospatial vegetation states.

We performed inferences on sampled data in the EarthNet2021X test dataset using the trained ConvLSTM, Earthformer, and VegeDiff models. Based on the inference results, we calculated the NDVI and ARVI indices, and displayed their images for the 5th, 10th, 15th, 20th, 25th, and 30th days, as illustrated in Figures \ref{Fig:ndvi_result} and Figure \ref{Fig:arvi_result}. Due to the deterministic nature of ConvLSTM and Earthformer, their forecastings tend to be blurry, which is disadvantageous for forecasting future vegetation states. Earthformer, in particular, produces even blurrier results as it divides images into several patches for forecasting. In contrast, VegeDiff employs a diffusion model to model the process of vegetation change. As a probabilistic model, it effectively handles the uncertainties in vegetation change, generating clear and accurate forecasting of vegetation states.

\subsection{Model Performance over Lead Time}

\begin{figure*}[!ht]
	\centering
		\includegraphics[width=0.95\linewidth]{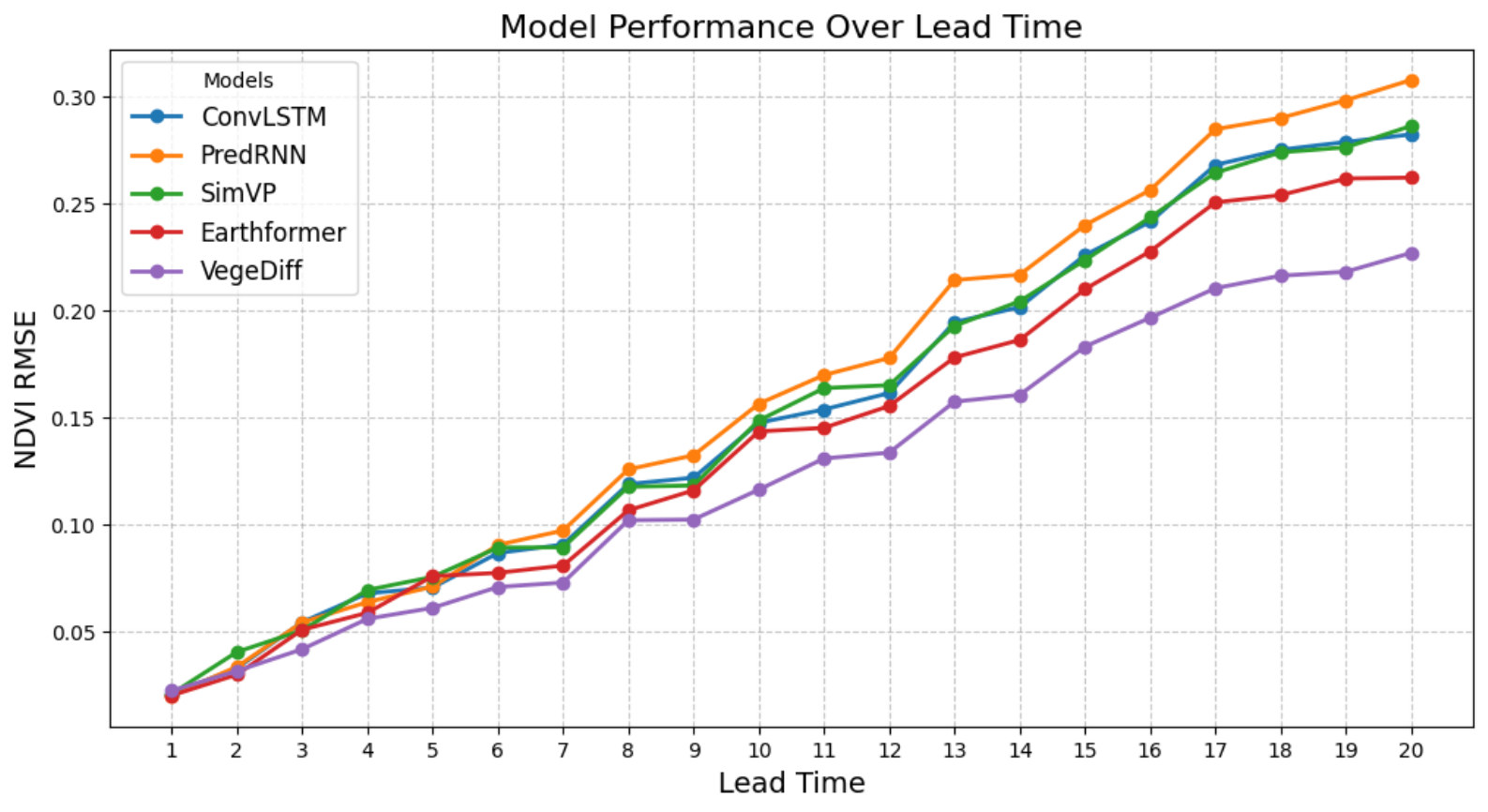}
	\caption{Illustration of model performance over lead time. Lead time refers to the interval between the forecasting time and the current time. Each lead time corresponds to a 5-day interval.}
 \label{Fig:lead_time}
\end{figure*}

Due to the complex influence of dynamic meteorological variables and static environmental variables, the geospatial vegetation change process exhibits a high degree of complexity and variability. Lead time refers to the interval between the forecasting time and the current time. Therefore, as the lead time of geospatial vegetation forecasting increases, both the degree and uncertainty of geospatial vegetation change increase, posing great challenges for vegetation forecasting. To demonstrate the superiority of VegeDiff in forecasting geospatial vegetation states over long lead time, we tested the performance of VegeDiff and comparative benchmark methods under different lead time. The model performance was evaluated by the RMSE between the predicted NDVI images and the ground truth, with lower RMSE indicating better model performance. The experimental results are shown in Figure \ref{Fig:lead_time}, where the horizontal axis represents the time interval between the current time and the forecasting time. It can be seen that as the lead time increases, the NDVI RMSE of all models increases. This is because the longer the lead time, the longer the geospatial vegetation is influenced by other variables, leading to greater degrees of change and uncertainty in geospatial vegetation, thereby increasing the difficulty of forecasting future vegetation states. Additionally, it is observed that when the lead time exceeds 3, VegeDiff outperforms all other models. In the case of forecasting future geospatial vegetation states over long lead time, VegeDiff significantly outperforms all comparison models, demonstrating its superiority in long lead time geospatial vegetation forecasting.

\subsection{Influence of Meteorological Variables}

\begin{table*}[!ht]
\centering
\caption{The influence of precipitation changes to the model performance on the EarthNet2021X test dataset. The best values are highlighted in bold.}
\label{precipitation_performance}
\begin{tabular}{l|cc|cc|cc}
\toprule
\multirow{2}{*}{Precipitation} & \multicolumn{2}{c|}{RGBN} & \multicolumn{2}{c|}{NDVI} & \multicolumn{2}{c}{ARVI} \\
 & RMSE $\downarrow$ & SSIM $\uparrow$ & RMSE $\downarrow$ & SSIM $\uparrow$ & RMSE $\downarrow$ & SSIM $\uparrow$ \\
\midrule
80\% & 0.08 & 0.87 & 0.21 & 0.86 & 0.20 & 0.77 \\
90\% & 0.06 & 0.92 & 0.16 & 0.87 & 0.17 & 0.81 \\
100\% & \textbf{0.04} & \textbf{0.94} & \textbf{0.13} & \textbf{0.89} & \textbf{0.14} & \textbf{0.82} \\
110\% & 0.05 & 0.93 & 0.15 & 0.87 & 0.15 & 0.80 \\
120\% & 0.08 & 0.88 & 0.20 & 0.84 & 0.18 & 0.76 \\
\bottomrule
\end{tabular}
\end{table*}

\begin{figure}[!ht]
	\centering
		\includegraphics[width=\linewidth]{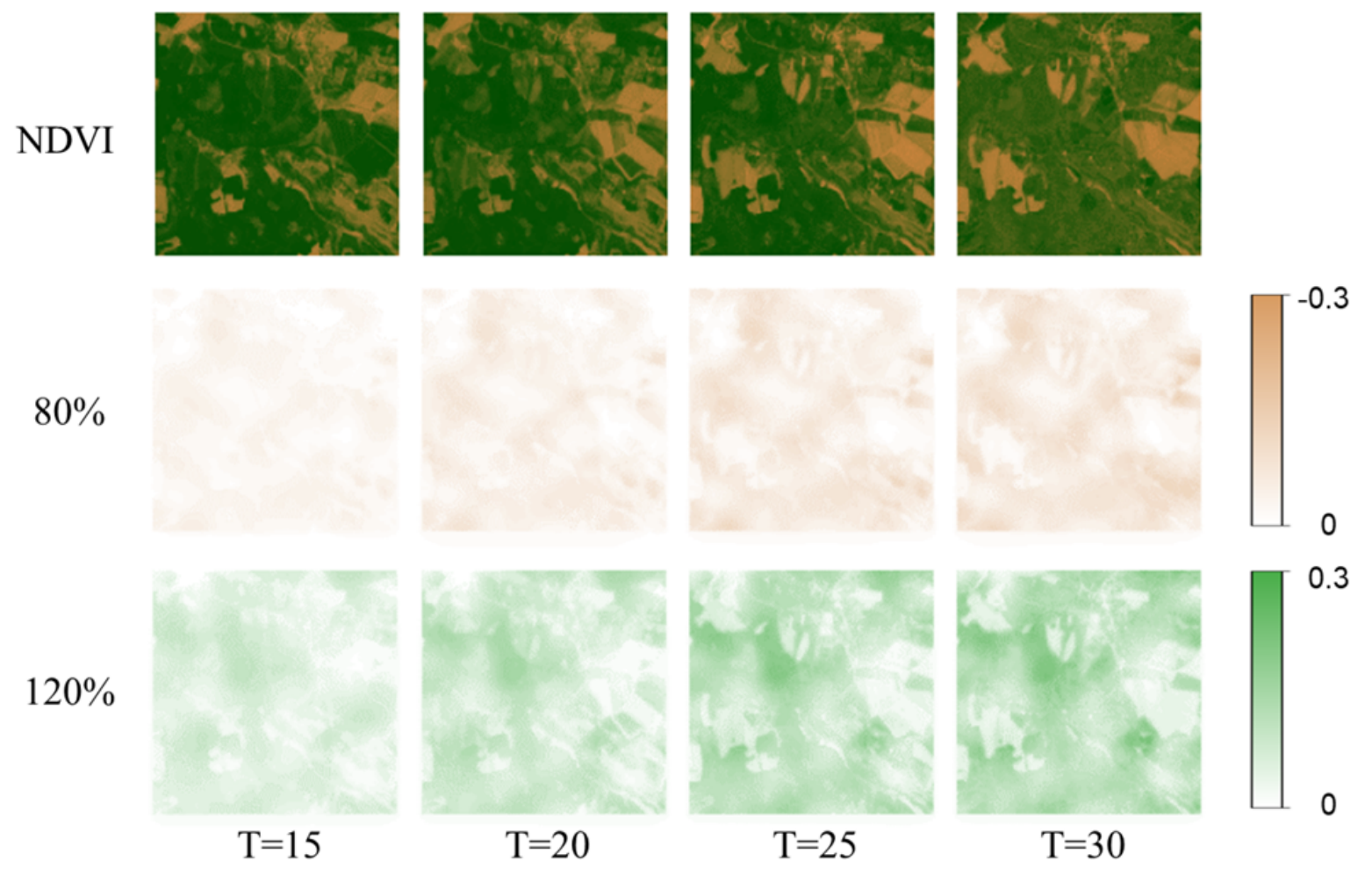}
	\caption{The influence of precipitation changes on geospatial vegetation changes. The first row represents the actual state of geospatial vegetation NDVI. The second row shows the difference between the NDVI when precipitation is reduced to 80\% of its original value and the actual NDVI. The third row displays the difference between the NDVI when precipitation is increased to 120\% of its original value and the actual NDVI.}
 \label{Fig:influence}
\end{figure}

Since VegeDiff can forecast future vegetation states under the constraints of dynamic meteorological variables and static environmental variables, we can use VegeDiff to explore the effects of these variables on geospatial vegetation. Specifically, we can modify one or more dynamic meteorological variables or static environmental variables, input the modified variables into VegeDiff, and compare the forecasting results with those obtained using the original variables. This approach helps us understand the influence of different dynamic meteorological variables and static environmental variables on vegetation changes, as well as the response of vegetation in different regions to these variable changes.

Given the significant impact of precipitation on vegetation changes, we focused on exploring its effect. We adjust precipitation to 80\%, 90\%, 100\% (no change), 110\%, and 120\% of the original values and test the model's performance under these conditions. As shown in Table \ref{precipitation_performance}, VegeDiff has the best vegetation states forecasting performance when precipitation remains unchanged (100\%). Both increased and decreased precipitation lead to erroneous forecastings, which is similar to the findings of~\citep{earthnet2021x_cvpr2021}. The results show that VegeDiff is sensitive to the change of precipitation and can accurately model the impact of precipitation on vegetation changes.

We displayed the NDVI deviations produced by VegeDiff under 80\% and 120\% precipitation conditions, as illustrated in Figure \ref{Fig:influence}. The first row shows the NDVI images on the 15th, 20th, 25th, and 30th days for a sample from the test set of the EarthNet2021X dataset. The second row shows the difference between the forecasting NDVI and the ground truth under 80\% precipitation, while the third row shows the difference under 120\% precipitation. It shows that VegeDiff underestimates NDVI when precipitation is at 80\% and overestimates NDVI at 120\%. As the forecasting period increases, the underestimation in the 80\% precipitation scenario becomes more pronounced, and the overestimation in the 120\% scenario also intensifies. This indicates that reduced precipitation inhibits vegetation growth, while increased precipitation promotes it, which is consistent with the findings of~\citep{earthnet2021X_arxiv}. Additionally, cumulative effects on vegetation NDVI arise from sustained changes in precipitation over time, with decreased or increased precipitation further reducing or enhancing NDVI, respectively. Moreover, different regions exhibit varied responses to changes in precipitation. Generally, areas with lush vegetation are more sensitive to precipitation changes and respond more strongly.

\section{Discussion}

\subsection{Forecasting Vegetation State from Multiple Perspectives}

\begin{figure*}[!ht]
	\centering
		\includegraphics[width=\linewidth]{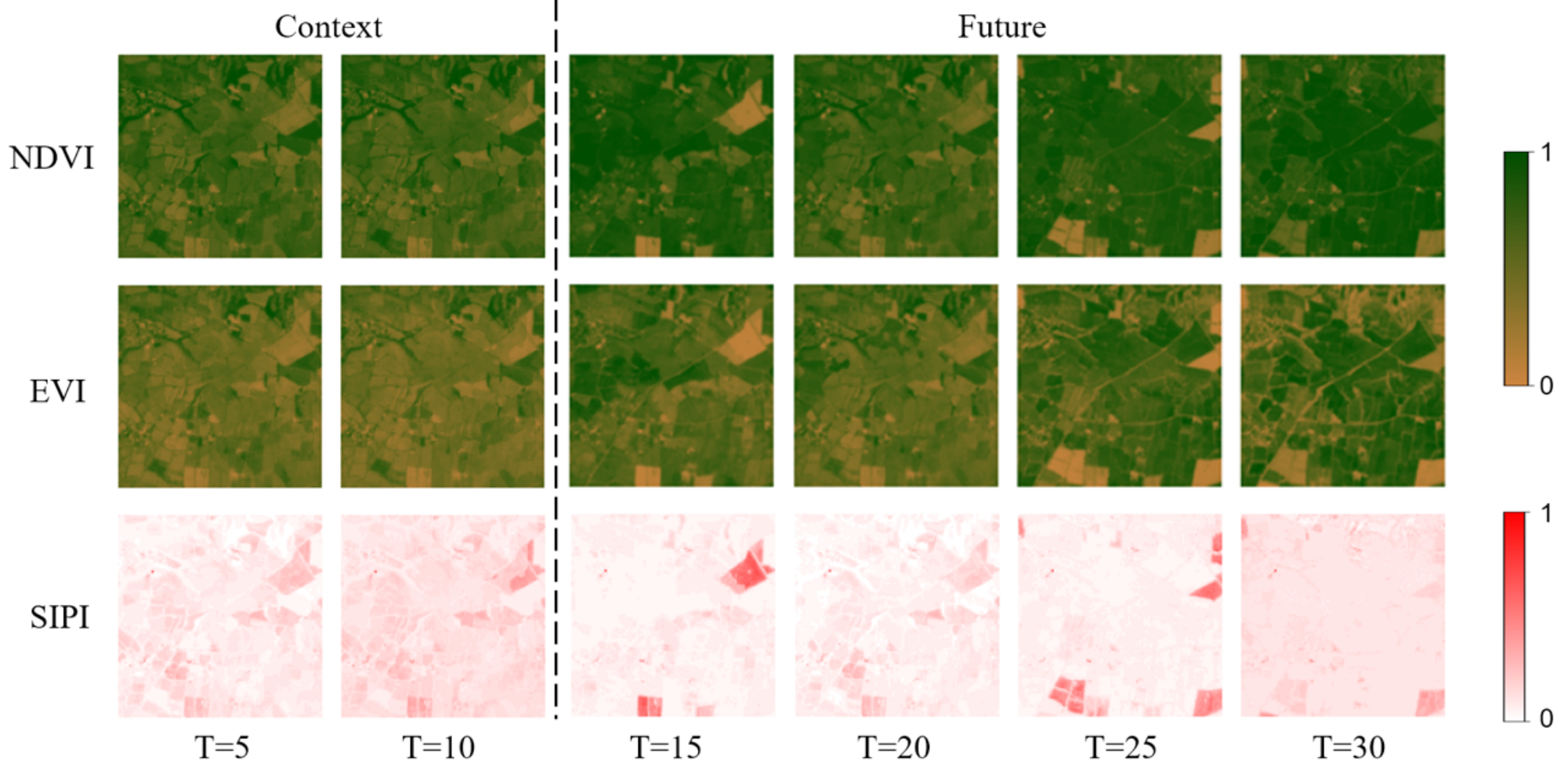}
	\caption{NDVI, EVI and SIPI forecasting results of VegeDiff on the EarthNet2021X dataset.}
 \label{Fig:multi_vege_index}
\end{figure*}

VegeDiff effectively forecasts future geospatial vegetation states based on past vegetation states, dynamic meteorological variables, and static environmental variables. By forecasting geospatial vegetation states with RGBN remote sensing images, VegeDiff can compute a wide range of vegetation indices from the forecasting results, providing comprehensive forecasting of future vegetation states. To demonstrate that VegeDiff can forecast the future state of geospatial vegetation from multiple perspectives, we selected three vegetation indices including NDVI, Enhanced Vegetation Index (EVI), and Structure Insensitive Pigment Index (SIPI), and displayed their forecasting results for the 15th, 20th, 25th, and 30th days, as shown in Figure \ref{Fig:multi_vege_index}.

NDVI is an effective indicator of vegetation health and density, primarily used to assess biomass and monitor vegetation changes over time. EVI offers improved sensitivity in high biomass regions and minimizes atmospheric and canopy background influences. SIPI is less sensitive to chlorophyll content variations and more indicative of the structure and condition of vegetation canopies.

As shown in Figure \ref{Fig:multi_vege_index}, VegeDiff can forecast geospatial vegetation states from multiple perspectives. The forecasting NDVI results (the first row) provides an overall indication of vegetation growth and effectively reflects vegetation health. The forecasting EVI results (the second row) corrects for soil and atmospheric effects, offering a more accurate representation of vegetation health and providing more detailed information on vegetation growth conditions. The forecasting SIPI results (the third row) effectively highlights areas with poor vegetation growth, serving as an early warning for vegetation diseases, geospatial drought, and other issues. Therefore, by forecasting RGBN remote sensing images through VegeDiff, we can compute various vegetation indices to forecast future vegetation states from multiple aspects. This capability provides valuable insights for agricultural management, disaster warning, and other applications.

\subsection{Expectations and Limitations}

The process of geospatial vegetation change is influenced by the complex interaction of dynamic meteorological variables and static environmental variables, resulting in a high degree of uncertainty. Current deterministic methods struggle to handle this uncertainty, often yielding vague and inaccurate forecasting results. Additionally, these methods do not adequately address the combined effects of dynamic meteorological and static environmental variables on geospatial vegetation, making them ineffective for vegetation forecasting tasks.

To address these issues, we propose VegeDiff for geospatial vegetation forecasting. VegeDiff utilizes a diffusion model to capture the uncertainties in vegetation change processes, enabling it to generate clear and accurate forecasting results. VegeNet within VegeDiff models the impacts of dynamic meteorological and static environmental variables on past vegetation states, facilitating accurate forecasting of future vegetation states. Ablation studies and comparative experiments on the EarthNet2021X dataset demonstrate the effectiveness of VegeDiff, which outperforms existing deterministic methods in forecasting geospatial vegetation states. What's more, VegeDiff significantly outperforms all comparison models over long lead time, showing its superiority in long lead time geospatial vegetation forecasting. 

Exploratory experiments on the EarthNet2021X dataset demonstrate the potential application value of VegeDiff. By generating multiple vegetation indices from RGBN forecasting results, VegeDiff can forecast future vegetation states from various perspectives. Furthermore, by altering one or more dynamic meteorological or static environmental variables, VegeDiff can forecast vegetation responses to changes in these variables, highlighting its potential to explore the effects of these variables on vegetation and different regional vegetation responses. We envision VegeDiff as a baseline in the field of geospatial vegetation forecasting, further promoting the development of probabilistic models in this domain, and uncovering its potential applications in practical scenarios. 

Despite its strong performance, VegeDiff also has several limitations. The pre-training and fine-tuning of the variational autoencoder in VegeDiff require substantial data and computation, and training VegeNet demands significant memory resources. These requirements pose challenges for efficiently transferring VegeDiff to other tasks. Additionally, as a latent diffusion model, forecasting large-scale and long-term future vegetation states with VegeDiff is time and resource-intensive, hindering its deployment and application in practical scenarios.

In the future, we aim to explore efficient fine-tuning techniques for VegeDiff, enabling its low-cost and high-efficiency transfer to other tasks. We also plan to investigate ways to reduce the denoising steps and accelerate the denoising process in VegeDiff, speeding up the vegetation state forecasting process and facilitating its deployment and application in practical scenarios.

\section{Conclusion}

We proposed VegeDiff for the geospatial vegetation forecasting task, which is based on the latent diffusion model. Current deterministic methods struggle with the inherent uncertainty in vegetation changes, often resulting in blurry and inaccurate forecasting results. VegeDiff employs a diffusion model to effectively capture this uncertainty and utilizes VegeNet to model the complex interactions between dynamic meteorological variables and static environmental variables, thereby accurately forecasting future vegetation states. The variational autoencoder within VegeDiff is pre-trained on 10M RGBN remote sensing data and fine-tuned on 20K vegetation remote sensing data. This extensive training process ensures a well-represented latent space, enhancing the model's understanding and predictive capability regarding vegetation states. 

Comparative experiments on the EarthNet2021X dataset demonstrate the effectiveness of VegeDiff in geospatial vegetation forecasting tasks. Unlike various deterministic methods, VegeDiff probabilistically models the uncertainty in the vegetation change process and separately models the impact of dynamic meteorological and static environmental variables on vegetation states, enabling it to generate clear and accurate forecasting results. We anticipate that VegeDiff will serve as a baseline in the field of vegetation forecasting, promoting the development of probabilistic models in this field and facilitating the deployment and application of VegeDiff in practical scenarios.

\section*{Acknowledgment}

This work was done during the internship of Sijie Zhao at Shanghai Artificial Intelligence Laboratory. This work is partially supported by the National Key R\&D Program of China (NO.2022ZD0160101), in part by the National Natural Science Foundation of China under Grant 42071297, and in part by the AI \& AI for Science Project of Nanjing University under Grant 02091480605203. The authors would like thank The Youth Innovation Team of China Meteorological Administration (CMA2024QN02) for their invaluable support and guidance throughout this research.


\printcredits



\begin{thebibliography}{56}
\expandafter\ifx\csname natexlab\endcsname\relax\def\natexlab#1{#1}\fi
\providecommand{\url}[1]{\texttt{#1}}
\providecommand{\href}[2]{#2}
\providecommand{\path}[1]{#1}
\providecommand{\DOIprefix}{doi:}
\providecommand{\ArXivprefix}{arXiv:}
\providecommand{\URLprefix}{URL: }
\providecommand{\Pubmedprefix}{pmid:}
\providecommand{\doi}[1]{\href{http://dx.doi.org/#1}{\path{#1}}}
\providecommand{\Pubmed}[1]{\href{pmid:#1}{\path{#1}}}
\providecommand{\bibinfo}[2]{#2}
\ifx\xfnm\relax \def\xfnm[#1]{\unskip,\space#1}\fi
\bibitem[{Asperti et~al.(2023)Asperti, Merizzi, Paparella, Pedrazzi, Angelinelli and Colamonaco}]{earth_diffusion1}
\bibinfo{author}{Asperti, A.}, \bibinfo{author}{Merizzi, F.}, \bibinfo{author}{Paparella, A.}, \bibinfo{author}{Pedrazzi, G.}, \bibinfo{author}{Angelinelli, M.}, \bibinfo{author}{Colamonaco, S.}, \bibinfo{year}{2023}.
\newblock \bibinfo{title}{Precipitation nowcasting with generative diffusion models}.
\newblock \bibinfo{journal}{arXiv preprint arXiv:2308.06733} .
\bibitem[{Barrett et~al.(2020)Barrett, Duivenvoorden, Salakpi, Muthoka, Mwangi, Oliver and Rowhani}]{vegetation2}
\bibinfo{author}{Barrett, A.B.}, \bibinfo{author}{Duivenvoorden, S.}, \bibinfo{author}{Salakpi, E.E.}, \bibinfo{author}{Muthoka, J.M.}, \bibinfo{author}{Mwangi, J.}, \bibinfo{author}{Oliver, S.}, \bibinfo{author}{Rowhani, P.}, \bibinfo{year}{2020}.
\newblock \bibinfo{title}{Forecasting vegetation condition for drought early warning systems in pastoral communities in kenya}.
\newblock \bibinfo{journal}{Remote Sensing of Environment} \bibinfo{volume}{248}, \bibinfo{pages}{111886}.
\bibitem[{Bastani et~al.(2023)Bastani, Wolters, Gupta, Ferdinando and Kembhavi}]{satlas}
\bibinfo{author}{Bastani, F.}, \bibinfo{author}{Wolters, P.}, \bibinfo{author}{Gupta, R.}, \bibinfo{author}{Ferdinando, J.}, \bibinfo{author}{Kembhavi, A.}, \bibinfo{year}{2023}.
\newblock \bibinfo{title}{Satlaspretrain: A large-scale dataset for remote sensing image understanding}, in: \bibinfo{booktitle}{Proceedings of the IEEE/CVF International Conference on Computer Vision}, pp. \bibinfo{pages}{16772--16782}.
\bibitem[{Bellprat et~al.(2019)Bellprat, Guemas, Doblas-Reyes and Donat}]{climate_change3}
\bibinfo{author}{Bellprat, O.}, \bibinfo{author}{Guemas, V.}, \bibinfo{author}{Doblas-Reyes, F.}, \bibinfo{author}{Donat, M.G.}, \bibinfo{year}{2019}.
\newblock \bibinfo{title}{Towards reliable extreme weather and climate event attribution}.
\newblock \bibinfo{journal}{Nature communications} \bibinfo{volume}{10}, \bibinfo{pages}{1732}.
\bibitem[{Benson et~al.(2023)Benson, Requena-Mesa, Robin, Alonso, Cort{\'e}s, Gao, Linscheid, Weynants and Reichstein}]{earthnet2021x}
\bibinfo{author}{Benson, V.}, \bibinfo{author}{Requena-Mesa, C.}, \bibinfo{author}{Robin, C.}, \bibinfo{author}{Alonso, L.}, \bibinfo{author}{Cort{\'e}s, J.}, \bibinfo{author}{Gao, Z.}, \bibinfo{author}{Linscheid, N.}, \bibinfo{author}{Weynants, M.}, \bibinfo{author}{Reichstein, M.}, \bibinfo{year}{2023}.
\newblock \bibinfo{title}{Forecasting localized weather impacts on vegetation as seen from space with meteo-guided video prediction}.
\newblock \bibinfo{journal}{arXiv preprint arXiv:2303.16198} .
\bibitem[{Benson et~al.(2024)Benson, Robin, Requena-Mesa, Alonso, Carvalhais, Cort{\'e}s, Gao, Linscheid, Weynants and Reichstein}]{earthnet2021x_cvpr2024}
\bibinfo{author}{Benson, V.}, \bibinfo{author}{Robin, C.}, \bibinfo{author}{Requena-Mesa, C.}, \bibinfo{author}{Alonso, L.}, \bibinfo{author}{Carvalhais, N.}, \bibinfo{author}{Cort{\'e}s, J.}, \bibinfo{author}{Gao, Z.}, \bibinfo{author}{Linscheid, N.}, \bibinfo{author}{Weynants, M.}, \bibinfo{author}{Reichstein, M.}, \bibinfo{year}{2024}.
\newblock \bibinfo{title}{Multi-modal learning for geospatial vegetation forecasting}, in: \bibinfo{booktitle}{Conference on Computer Vision and Pattern Recognition 2024}.
\bibitem[{Bi et~al.(2023)Bi, Xie, Zhang, Chen, Gu and Tian}]{big_data_dl3}
\bibinfo{author}{Bi, K.}, \bibinfo{author}{Xie, L.}, \bibinfo{author}{Zhang, H.}, \bibinfo{author}{Chen, X.}, \bibinfo{author}{Gu, X.}, \bibinfo{author}{Tian, Q.}, \bibinfo{year}{2023}.
\newblock \bibinfo{title}{Accurate medium-range global weather forecasting with 3d neural networks}.
\newblock \bibinfo{journal}{Nature} \bibinfo{volume}{619}, \bibinfo{pages}{533--538}.
\bibitem[{Blattmann et~al.(2023)Blattmann, Rombach, Ling, Dockhorn, Kim, Fidler and Kreis}]{ldm_video_work1}
\bibinfo{author}{Blattmann, A.}, \bibinfo{author}{Rombach, R.}, \bibinfo{author}{Ling, H.}, \bibinfo{author}{Dockhorn, T.}, \bibinfo{author}{Kim, S.W.}, \bibinfo{author}{Fidler, S.}, \bibinfo{author}{Kreis, K.}, \bibinfo{year}{2023}.
\newblock \bibinfo{title}{Align your latents: High-resolution video synthesis with latent diffusion models}, in: \bibinfo{booktitle}{Proceedings of the IEEE/CVF Conference on Computer Vision and Pattern Recognition}, pp. \bibinfo{pages}{22563--22575}.
\bibitem[{Chen et~al.(2023)Chen, Sun, Song and Luo}]{dm_work1}
\bibinfo{author}{Chen, S.}, \bibinfo{author}{Sun, P.}, \bibinfo{author}{Song, Y.}, \bibinfo{author}{Luo, P.}, \bibinfo{year}{2023}.
\newblock \bibinfo{title}{Diffusiondet: Diffusion model for object detection}, in: \bibinfo{booktitle}{Proceedings of the IEEE/CVF International Conference on Computer Vision}, pp. \bibinfo{pages}{19830--19843}.
\bibitem[{Cornes et~al.(2018)Cornes, van~der Schrier, van~den Besselaar and Jones}]{eobs}
\bibinfo{author}{Cornes, R.C.}, \bibinfo{author}{van~der Schrier, G.}, \bibinfo{author}{van~den Besselaar, E.J.}, \bibinfo{author}{Jones, P.D.}, \bibinfo{year}{2018}.
\newblock \bibinfo{title}{An ensemble version of the e-obs temperature and precipitation data sets}.
\newblock \bibinfo{journal}{Journal of Geophysical Research: Atmospheres} \bibinfo{volume}{123}, \bibinfo{pages}{9391--9409}.
\bibitem[{Danier et~al.(2024)Danier, Zhang and Bull}]{ldm_video_work2}
\bibinfo{author}{Danier, D.}, \bibinfo{author}{Zhang, F.}, \bibinfo{author}{Bull, D.}, \bibinfo{year}{2024}.
\newblock \bibinfo{title}{Ldmvfi: Video frame interpolation with latent diffusion models}, in: \bibinfo{booktitle}{Proceedings of the AAAI Conference on Artificial Intelligence}, pp. \bibinfo{pages}{1472--1480}.
\bibitem[{Deser et~al.(2020)Deser, Lehner, Rodgers, Ault, Delworth, DiNezio, Fiore, Frankignoul, Fyfe, Horton et~al.}]{uncertainty1}
\bibinfo{author}{Deser, C.}, \bibinfo{author}{Lehner, F.}, \bibinfo{author}{Rodgers, K.B.}, \bibinfo{author}{Ault, T.}, \bibinfo{author}{Delworth, T.L.}, \bibinfo{author}{DiNezio, P.N.}, \bibinfo{author}{Fiore, A.}, \bibinfo{author}{Frankignoul, C.}, \bibinfo{author}{Fyfe, J.C.}, \bibinfo{author}{Horton, D.E.}, et~al., \bibinfo{year}{2020}.
\newblock \bibinfo{title}{Insights from earth system model initial-condition large ensembles and future prospects}.
\newblock \bibinfo{journal}{Nature Climate Change} \bibinfo{volume}{10}, \bibinfo{pages}{277--286}.
\bibitem[{Diaconu et~al.(2022)Diaconu, Saha, G{\"u}nnemann and Zhu}]{earthnet2021x_cvpr2021}
\bibinfo{author}{Diaconu, C.A.}, \bibinfo{author}{Saha, S.}, \bibinfo{author}{G{\"u}nnemann, S.}, \bibinfo{author}{Zhu, X.X.}, \bibinfo{year}{2022}.
\newblock \bibinfo{title}{Understanding the role of weather data for earth surface forecasting using a convlstm-based model}, in: \bibinfo{booktitle}{Proceedings of the IEEE/CVF Conference on Computer Vision and Pattern Recognition}, pp. \bibinfo{pages}{1362--1371}.
\bibitem[{Fathi et~al.(2022)Fathi, Haghi~Kashani, Jameii and Mahdipour}]{big_data_me}
\bibinfo{author}{Fathi, M.}, \bibinfo{author}{Haghi~Kashani, M.}, \bibinfo{author}{Jameii, S.M.}, \bibinfo{author}{Mahdipour, E.}, \bibinfo{year}{2022}.
\newblock \bibinfo{title}{Big data analytics in weather forecasting: A systematic review}.
\newblock \bibinfo{journal}{Archives of Computational Methods in Engineering} \bibinfo{volume}{29}, \bibinfo{pages}{1247--1275}.
\bibitem[{Fernandez et~al.(2023)Fernandez, Couairon, J{\'e}gou, Douze and Furon}]{ldm_work3}
\bibinfo{author}{Fernandez, P.}, \bibinfo{author}{Couairon, G.}, \bibinfo{author}{J{\'e}gou, H.}, \bibinfo{author}{Douze, M.}, \bibinfo{author}{Furon, T.}, \bibinfo{year}{2023}.
\newblock \bibinfo{title}{The stable signature: Rooting watermarks in latent diffusion models}, in: \bibinfo{booktitle}{Proceedings of the IEEE/CVF International Conference on Computer Vision}, pp. \bibinfo{pages}{22466--22477}.
\bibitem[{Gao et~al.(2024)Gao, Shi, Han, Wang, Jin, Maddix, Zhu, Li and Wang}]{prediff}
\bibinfo{author}{Gao, Z.}, \bibinfo{author}{Shi, X.}, \bibinfo{author}{Han, B.}, \bibinfo{author}{Wang, H.}, \bibinfo{author}{Jin, X.}, \bibinfo{author}{Maddix, D.}, \bibinfo{author}{Zhu, Y.}, \bibinfo{author}{Li, M.}, \bibinfo{author}{Wang, Y.B.}, \bibinfo{year}{2024}.
\newblock \bibinfo{title}{Prediff: Precipitation nowcasting with latent diffusion models}.
\newblock \bibinfo{journal}{Advances in Neural Information Processing Systems} \bibinfo{volume}{36}.
\bibitem[{Gao et~al.(2022)Gao, Shi, Wang, Zhu, Wang, Li and Yeung}]{earthformer}
\bibinfo{author}{Gao, Z.}, \bibinfo{author}{Shi, X.}, \bibinfo{author}{Wang, H.}, \bibinfo{author}{Zhu, Y.}, \bibinfo{author}{Wang, Y.B.}, \bibinfo{author}{Li, M.}, \bibinfo{author}{Yeung, D.Y.}, \bibinfo{year}{2022}.
\newblock \bibinfo{title}{Earthformer: Exploring space-time transformers for earth system forecasting}.
\newblock \bibinfo{journal}{Advances in Neural Information Processing Systems} \bibinfo{volume}{35}, \bibinfo{pages}{25390--25403}.
\bibitem[{Goyal et~al.(2017)Goyal, Doll{\'a}r, Girshick, Noordhuis, Wesolowski, Kyrola, Tulloch, Jia and He}]{adaln2}
\bibinfo{author}{Goyal, P.}, \bibinfo{author}{Doll{\'a}r, P.}, \bibinfo{author}{Girshick, R.}, \bibinfo{author}{Noordhuis, P.}, \bibinfo{author}{Wesolowski, L.}, \bibinfo{author}{Kyrola, A.}, \bibinfo{author}{Tulloch, A.}, \bibinfo{author}{Jia, Y.}, \bibinfo{author}{He, K.}, \bibinfo{year}{2017}.
\newblock \bibinfo{title}{Accurate, large minibatch sgd: Training imagenet in 1 hour}.
\newblock \bibinfo{journal}{arXiv preprint arXiv:1706.02677} .
\bibitem[{Ho et~al.(2020)Ho, Jain and Abbeel}]{dm}
\bibinfo{author}{Ho, J.}, \bibinfo{author}{Jain, A.}, \bibinfo{author}{Abbeel, P.}, \bibinfo{year}{2020}.
\newblock \bibinfo{title}{Denoising diffusion probabilistic models}.
\newblock \bibinfo{journal}{Advances in neural information processing systems} \bibinfo{volume}{33}, \bibinfo{pages}{6840--6851}.
\bibitem[{Ho et~al.(2022)Ho, Salimans, Gritsenko, Chan, Norouzi and Fleet}]{dm_work3}
\bibinfo{author}{Ho, J.}, \bibinfo{author}{Salimans, T.}, \bibinfo{author}{Gritsenko, A.}, \bibinfo{author}{Chan, W.}, \bibinfo{author}{Norouzi, M.}, \bibinfo{author}{Fleet, D.J.}, \bibinfo{year}{2022}.
\newblock \bibinfo{title}{Video diffusion models}.
\newblock \bibinfo{journal}{Advances in Neural Information Processing Systems} \bibinfo{volume}{35}, \bibinfo{pages}{8633--8646}.
\bibitem[{Kingma et~al.(2021)Kingma, Salimans, Poole and Ho}]{dm_work2}
\bibinfo{author}{Kingma, D.}, \bibinfo{author}{Salimans, T.}, \bibinfo{author}{Poole, B.}, \bibinfo{author}{Ho, J.}, \bibinfo{year}{2021}.
\newblock \bibinfo{title}{Variational diffusion models}.
\newblock \bibinfo{journal}{Advances in neural information processing systems} \bibinfo{volume}{34}, \bibinfo{pages}{21696--21707}.
\bibitem[{Kingma and Ba(2014)}]{adam}
\bibinfo{author}{Kingma, D.P.}, \bibinfo{author}{Ba, J.}, \bibinfo{year}{2014}.
\newblock \bibinfo{title}{Adam: A method for stochastic optimization}.
\newblock \bibinfo{journal}{arXiv preprint arXiv:1412.6980} .
\bibitem[{Kingma and Welling(2013)}]{vae}
\bibinfo{author}{Kingma, D.P.}, \bibinfo{author}{Welling, M.}, \bibinfo{year}{2013}.
\newblock \bibinfo{title}{Auto-encoding variational bayes}.
\newblock \bibinfo{journal}{arXiv preprint arXiv:1312.6114} .
\bibitem[{Kladny et~al.(2022)Kladny, Milanta, Mraz, Hufkens and Stocker}]{earthnet2021X_arxiv}
\bibinfo{author}{Kladny, K.R.}, \bibinfo{author}{Milanta, M.}, \bibinfo{author}{Mraz, O.}, \bibinfo{author}{Hufkens, K.}, \bibinfo{author}{Stocker, B.D.}, \bibinfo{year}{2022}.
\newblock \bibinfo{title}{Deep learning for satellite image forecasting of vegetation greenness}.
\newblock \bibinfo{journal}{bioRxiv} , \bibinfo{pages}{2022--08}.
\bibitem[{Leinonen et~al.(2023)Leinonen, Hamann, Nerini, Germann and Franch}]{ldcast}
\bibinfo{author}{Leinonen, J.}, \bibinfo{author}{Hamann, U.}, \bibinfo{author}{Nerini, D.}, \bibinfo{author}{Germann, U.}, \bibinfo{author}{Franch, G.}, \bibinfo{year}{2023}.
\newblock \bibinfo{title}{Latent diffusion models for generative precipitation nowcasting with accurate uncertainty quantification}.
\newblock \bibinfo{journal}{arXiv preprint arXiv:2304.12891} .
\bibitem[{Li et~al.(2024)Li, Carver, Lopez-Gomez, Sha and Anderson}]{earth_diffusion2}
\bibinfo{author}{Li, L.}, \bibinfo{author}{Carver, R.}, \bibinfo{author}{Lopez-Gomez, I.}, \bibinfo{author}{Sha, F.}, \bibinfo{author}{Anderson, J.}, \bibinfo{year}{2024}.
\newblock \bibinfo{title}{Generative emulation of weather forecast ensembles with diffusion models}.
\newblock \bibinfo{journal}{Science Advances} \bibinfo{volume}{10}, \bibinfo{pages}{eadk4489}.
\bibitem[{Li et~al.(2023)Li, Feng, Ran, Su, Liu, Huang, Shen, Xiao, Su, Yuan et~al.}]{big_data_dl2}
\bibinfo{author}{Li, X.}, \bibinfo{author}{Feng, M.}, \bibinfo{author}{Ran, Y.}, \bibinfo{author}{Su, Y.}, \bibinfo{author}{Liu, F.}, \bibinfo{author}{Huang, C.}, \bibinfo{author}{Shen, H.}, \bibinfo{author}{Xiao, Q.}, \bibinfo{author}{Su, J.}, \bibinfo{author}{Yuan, S.}, et~al., \bibinfo{year}{2023}.
\newblock \bibinfo{title}{Big data in earth system science and progress towards a digital twin}.
\newblock \bibinfo{journal}{Nature Reviews Earth \& Environment} \bibinfo{volume}{4}, \bibinfo{pages}{319--332}.
\bibitem[{Louis et~al.(2016)Louis, Debaecker, Pflug, Main-Knorn, Bieniarz, Mueller-Wilm, Cadau and Gascon}]{sentinel}
\bibinfo{author}{Louis, J.}, \bibinfo{author}{Debaecker, V.}, \bibinfo{author}{Pflug, B.}, \bibinfo{author}{Main-Knorn, M.}, \bibinfo{author}{Bieniarz, J.}, \bibinfo{author}{Mueller-Wilm, U.}, \bibinfo{author}{Cadau, E.}, \bibinfo{author}{Gascon, F.}, \bibinfo{year}{2016}.
\newblock \bibinfo{title}{Sentinel-2 sen2cor: L2a processor for users}, in: \bibinfo{booktitle}{Proceedings living planet symposium 2016}, \bibinfo{organization}{Spacebooks Online}. pp. \bibinfo{pages}{1--8}.
\bibitem[{Luo et~al.(2023)Luo, Chen, Zhang, Huang, Wang, Shen, Zhao, Zhou and Tan}]{video_fusion}
\bibinfo{author}{Luo, Z.}, \bibinfo{author}{Chen, D.}, \bibinfo{author}{Zhang, Y.}, \bibinfo{author}{Huang, Y.}, \bibinfo{author}{Wang, L.}, \bibinfo{author}{Shen, Y.}, \bibinfo{author}{Zhao, D.}, \bibinfo{author}{Zhou, J.}, \bibinfo{author}{Tan, T.}, \bibinfo{year}{2023}.
\newblock \bibinfo{title}{Videofusion: Decomposed diffusion models for high-quality video generation}, in: \bibinfo{booktitle}{Proceedings of the IEEE/CVF Conference on Computer Vision and Pattern Recognition}, pp. \bibinfo{pages}{10209--10218}.
\bibitem[{Miao et~al.(2024)Miao, Ju, Sun, Agathokleous, Wang, Zhu, Liu, Zou, Lu and Liu}]{climate_change1}
\bibinfo{author}{Miao, L.}, \bibinfo{author}{Ju, L.}, \bibinfo{author}{Sun, S.}, \bibinfo{author}{Agathokleous, E.}, \bibinfo{author}{Wang, Q.}, \bibinfo{author}{Zhu, Z.}, \bibinfo{author}{Liu, R.}, \bibinfo{author}{Zou, Y.}, \bibinfo{author}{Lu, Y.}, \bibinfo{author}{Liu, Q.}, \bibinfo{year}{2024}.
\newblock \bibinfo{title}{Unveiling the dynamics of sequential extreme precipitation-heatwave compounds in china}.
\newblock \bibinfo{journal}{npj Climate and Atmospheric Science} \bibinfo{volume}{7}, \bibinfo{pages}{67}.
\bibitem[{Millet et~al.(2019)Millet, Kruijer, Coupel-Ledru, Alvarez~Prado, Cabrera-Bosquet, Lacube, Charcosset, Welcker, van Eeuwijk and Tardieu}]{crop_yield1}
\bibinfo{author}{Millet, E.J.}, \bibinfo{author}{Kruijer, W.}, \bibinfo{author}{Coupel-Ledru, A.}, \bibinfo{author}{Alvarez~Prado, S.}, \bibinfo{author}{Cabrera-Bosquet, L.}, \bibinfo{author}{Lacube, S.}, \bibinfo{author}{Charcosset, A.}, \bibinfo{author}{Welcker, C.}, \bibinfo{author}{van Eeuwijk, F.}, \bibinfo{author}{Tardieu, F.}, \bibinfo{year}{2019}.
\newblock \bibinfo{title}{Genomic prediction of maize yield across european environmental conditions}.
\newblock \bibinfo{journal}{Nature genetics} \bibinfo{volume}{51}, \bibinfo{pages}{952--956}.
\bibitem[{Ni et~al.(2023)Ni, Shi, Li, Huang and Min}]{lfdm}
\bibinfo{author}{Ni, H.}, \bibinfo{author}{Shi, C.}, \bibinfo{author}{Li, K.}, \bibinfo{author}{Huang, S.X.}, \bibinfo{author}{Min, M.R.}, \bibinfo{year}{2023}.
\newblock \bibinfo{title}{Conditional image-to-video generation with latent flow diffusion models}, in: \bibinfo{booktitle}{Proceedings of the IEEE/CVF Conference on Computer Vision and Pattern Recognition}, pp. \bibinfo{pages}{18444--18455}.
\bibitem[{Paszke et~al.(2017)Paszke, Gross, Chintala, Chanan, Yang, DeVito, Lin, Desmaison, Antiga and Lerer}]{pytorch}
\bibinfo{author}{Paszke, A.}, \bibinfo{author}{Gross, S.}, \bibinfo{author}{Chintala, S.}, \bibinfo{author}{Chanan, G.}, \bibinfo{author}{Yang, E.}, \bibinfo{author}{DeVito, Z.}, \bibinfo{author}{Lin, Z.}, \bibinfo{author}{Desmaison, A.}, \bibinfo{author}{Antiga, L.}, \bibinfo{author}{Lerer, A.}, \bibinfo{year}{2017}.
\newblock \bibinfo{title}{Automatic differentiation in pytorch} .
\bibitem[{Peebles and Xie(2023)}]{dit}
\bibinfo{author}{Peebles, W.}, \bibinfo{author}{Xie, S.}, \bibinfo{year}{2023}.
\newblock \bibinfo{title}{Scalable diffusion models with transformers}, in: \bibinfo{booktitle}{Proceedings of the IEEE/CVF International Conference on Computer Vision}, pp. \bibinfo{pages}{4195--4205}.
\bibitem[{Perez et~al.(2018)Perez, Strub, De~Vries, Dumoulin and Courville}]{adaln1}
\bibinfo{author}{Perez, E.}, \bibinfo{author}{Strub, F.}, \bibinfo{author}{De~Vries, H.}, \bibinfo{author}{Dumoulin, V.}, \bibinfo{author}{Courville, A.}, \bibinfo{year}{2018}.
\newblock \bibinfo{title}{Film: Visual reasoning with a general conditioning layer}, in: \bibinfo{booktitle}{Proceedings of the AAAI conference on artificial intelligence}.
\bibitem[{Pinaya et~al.(2022)Pinaya, Tudosiu, Dafflon, Da~Costa, Fernandez, Nachev, Ourselin and Cardoso}]{ldm_work2}
\bibinfo{author}{Pinaya, W.H.}, \bibinfo{author}{Tudosiu, P.D.}, \bibinfo{author}{Dafflon, J.}, \bibinfo{author}{Da~Costa, P.F.}, \bibinfo{author}{Fernandez, V.}, \bibinfo{author}{Nachev, P.}, \bibinfo{author}{Ourselin, S.}, \bibinfo{author}{Cardoso, M.J.}, \bibinfo{year}{2022}.
\newblock \bibinfo{title}{Brain imaging generation with latent diffusion models}, in: \bibinfo{booktitle}{MICCAI Workshop on Deep Generative Models}, \bibinfo{organization}{Springer}. pp. \bibinfo{pages}{117--126}.
\bibitem[{Reichstein et~al.(2019)Reichstein, Camps-Valls, Stevens, Jung, Denzler, Carvalhais and Prabhat}]{big_data_dl1}
\bibinfo{author}{Reichstein, M.}, \bibinfo{author}{Camps-Valls, G.}, \bibinfo{author}{Stevens, B.}, \bibinfo{author}{Jung, M.}, \bibinfo{author}{Denzler, J.}, \bibinfo{author}{Carvalhais, N.}, \bibinfo{author}{Prabhat, f.}, \bibinfo{year}{2019}.
\newblock \bibinfo{title}{Deep learning and process understanding for data-driven earth system science}.
\newblock \bibinfo{journal}{Nature} \bibinfo{volume}{566}, \bibinfo{pages}{195--204}.
\bibitem[{Requena-Mesa et~al.(2021)Requena-Mesa, Benson, Reichstein, Runge and Denzler}]{earthnet2021}
\bibinfo{author}{Requena-Mesa, C.}, \bibinfo{author}{Benson, V.}, \bibinfo{author}{Reichstein, M.}, \bibinfo{author}{Runge, J.}, \bibinfo{author}{Denzler, J.}, \bibinfo{year}{2021}.
\newblock \bibinfo{title}{Earthnet2021: A large-scale dataset and challenge for earth surface forecasting as a guided video prediction task.}, in: \bibinfo{booktitle}{Proceedings of the IEEE/CVF Conference on Computer Vision and Pattern Recognition}, pp. \bibinfo{pages}{1132--1142}.
\bibitem[{Robin et~al.(2022)Robin, Requena-Mesa, Benson, Alonso, Poehls, Carvalhais and Reichstein}]{africa_dataset}
\bibinfo{author}{Robin, C.}, \bibinfo{author}{Requena-Mesa, C.}, \bibinfo{author}{Benson, V.}, \bibinfo{author}{Alonso, L.}, \bibinfo{author}{Poehls, J.}, \bibinfo{author}{Carvalhais, N.}, \bibinfo{author}{Reichstein, M.}, \bibinfo{year}{2022}.
\newblock \bibinfo{title}{Learning to forecast vegetation greenness at fine resolution over africa with convlstms}.
\newblock \bibinfo{journal}{arXiv preprint arXiv:2210.13648} .
\bibitem[{Rombach et~al.(2022)Rombach, Blattmann, Lorenz, Esser and Ommer}]{ldm}
\bibinfo{author}{Rombach, R.}, \bibinfo{author}{Blattmann, A.}, \bibinfo{author}{Lorenz, D.}, \bibinfo{author}{Esser, P.}, \bibinfo{author}{Ommer, B.}, \bibinfo{year}{2022}.
\newblock \bibinfo{title}{High-resolution image synthesis with latent diffusion models}, in: \bibinfo{booktitle}{Proceedings of the IEEE/CVF conference on computer vision and pattern recognition}, pp. \bibinfo{pages}{10684--10695}.
\bibitem[{Sagan et~al.(2021)Sagan, Maimaitijiang, Bhadra, Maimaitiyiming, Brown, Sidike and Fritschi}]{crop_yield2}
\bibinfo{author}{Sagan, V.}, \bibinfo{author}{Maimaitijiang, M.}, \bibinfo{author}{Bhadra, S.}, \bibinfo{author}{Maimaitiyiming, M.}, \bibinfo{author}{Brown, D.R.}, \bibinfo{author}{Sidike, P.}, \bibinfo{author}{Fritschi, F.B.}, \bibinfo{year}{2021}.
\newblock \bibinfo{title}{Field-scale crop yield prediction using multi-temporal worldview-3 and planetscope satellite data and deep learning}.
\newblock \bibinfo{journal}{ISPRS journal of photogrammetry and remote sensing} \bibinfo{volume}{174}, \bibinfo{pages}{265--281}.
\bibitem[{Song et~al.(2018)Song, Hansen, Stehman, Potapov, Tyukavina, Vermote and Townshend}]{lulc_change1}
\bibinfo{author}{Song, X.P.}, \bibinfo{author}{Hansen, M.C.}, \bibinfo{author}{Stehman, S.V.}, \bibinfo{author}{Potapov, P.V.}, \bibinfo{author}{Tyukavina, A.}, \bibinfo{author}{Vermote, E.F.}, \bibinfo{author}{Townshend, J.R.}, \bibinfo{year}{2018}.
\newblock \bibinfo{title}{Global land change from 1982 to 2016}.
\newblock \bibinfo{journal}{Nature} \bibinfo{volume}{560}, \bibinfo{pages}{639--643}.
\bibitem[{Takagi and Nishimoto(2023)}]{ldm_work1}
\bibinfo{author}{Takagi, Y.}, \bibinfo{author}{Nishimoto, S.}, \bibinfo{year}{2023}.
\newblock \bibinfo{title}{High-resolution image reconstruction with latent diffusion models from human brain activity}, in: \bibinfo{booktitle}{Proceedings of the IEEE/CVF Conference on Computer Vision and Pattern Recognition}, pp. \bibinfo{pages}{14453--14463}.
\bibitem[{Tan et~al.(2022)Tan, Gao, Li and Li}]{simvp}
\bibinfo{author}{Tan, C.}, \bibinfo{author}{Gao, Z.}, \bibinfo{author}{Li, S.}, \bibinfo{author}{Li, S.Z.}, \bibinfo{year}{2022}.
\newblock \bibinfo{title}{Simvp: Towards simple yet powerful spatiotemporal predictive learning}.
\newblock \bibinfo{journal}{arXiv preprint arXiv:2211.12509} .
\bibitem[{Tian et~al.(2019)Tian, Van~Dijk, Tregoning and Renzullo}]{vegetation1}
\bibinfo{author}{Tian, S.}, \bibinfo{author}{Van~Dijk, A.I.}, \bibinfo{author}{Tregoning, P.}, \bibinfo{author}{Renzullo, L.J.}, \bibinfo{year}{2019}.
\newblock \bibinfo{title}{Forecasting dryland vegetation condition months in advance through satellite data assimilation}.
\newblock \bibinfo{journal}{Nature Communications} \bibinfo{volume}{10}, \bibinfo{pages}{469}.
\bibitem[{Trenberth et~al.(2015)Trenberth, Fasullo and Shepherd}]{climate_change4}
\bibinfo{author}{Trenberth, K.E.}, \bibinfo{author}{Fasullo, J.T.}, \bibinfo{author}{Shepherd, T.G.}, \bibinfo{year}{2015}.
\newblock \bibinfo{title}{Attribution of climate extreme events}.
\newblock \bibinfo{journal}{Nature Climate Change} \bibinfo{volume}{5}, \bibinfo{pages}{725--730}.
\bibitem[{Wang et~al.(2022a)Wang, Chen, Jarin and Xie}]{climate_change2}
\bibinfo{author}{Wang, D.}, \bibinfo{author}{Chen, Y.}, \bibinfo{author}{Jarin, M.}, \bibinfo{author}{Xie, X.}, \bibinfo{year}{2022}a.
\newblock \bibinfo{title}{Increasingly frequent extreme weather events urge the development of point-of-use water treatment systems}.
\newblock \bibinfo{journal}{npj Clean Water} \bibinfo{volume}{5}, \bibinfo{pages}{36}.
\bibitem[{Wang et~al.(2022b)Wang, Wu, Zhang, Gao, Wang, Philip and Long}]{predrnn}
\bibinfo{author}{Wang, Y.}, \bibinfo{author}{Wu, H.}, \bibinfo{author}{Zhang, J.}, \bibinfo{author}{Gao, Z.}, \bibinfo{author}{Wang, J.}, \bibinfo{author}{Philip, S.Y.}, \bibinfo{author}{Long, M.}, \bibinfo{year}{2022}b.
\newblock \bibinfo{title}{Predrnn: A recurrent neural network for spatiotemporal predictive learning}.
\newblock \bibinfo{journal}{IEEE Transactions on Pattern Analysis and Machine Intelligence} \bibinfo{volume}{45}, \bibinfo{pages}{2208--2225}.
\bibitem[{Wu et~al.(2015)Wu, Zhao, Liang, Zhou, Huang, Tang and Zhao}]{influence1}
\bibinfo{author}{Wu, D.}, \bibinfo{author}{Zhao, X.}, \bibinfo{author}{Liang, S.}, \bibinfo{author}{Zhou, T.}, \bibinfo{author}{Huang, K.}, \bibinfo{author}{Tang, B.}, \bibinfo{author}{Zhao, W.}, \bibinfo{year}{2015}.
\newblock \bibinfo{title}{Time-lag effects of global vegetation responses to climate change}.
\newblock \bibinfo{journal}{Global change biology} \bibinfo{volume}{21}, \bibinfo{pages}{3520--3531}.
\bibitem[{Xia et~al.(2023)Xia, Zhang, Wang, Wang, Wu, Tian, Yang and Van~Gool}]{dm_work4}
\bibinfo{author}{Xia, B.}, \bibinfo{author}{Zhang, Y.}, \bibinfo{author}{Wang, S.}, \bibinfo{author}{Wang, Y.}, \bibinfo{author}{Wu, X.}, \bibinfo{author}{Tian, Y.}, \bibinfo{author}{Yang, W.}, \bibinfo{author}{Van~Gool, L.}, \bibinfo{year}{2023}.
\newblock \bibinfo{title}{Diffir: Efficient diffusion model for image restoration}, in: \bibinfo{booktitle}{Proceedings of the IEEE/CVF International Conference on Computer Vision}, pp. \bibinfo{pages}{13095--13105}.
\bibitem[{Yan et~al.(2023)Yan, Sun, Eldardiry, Thurber, Reed, Malek, Gupta, Kennedy, Swenson, Wang et~al.}]{uncertainty2}
\bibinfo{author}{Yan, H.}, \bibinfo{author}{Sun, N.}, \bibinfo{author}{Eldardiry, H.}, \bibinfo{author}{Thurber, T.B.}, \bibinfo{author}{Reed, P.M.}, \bibinfo{author}{Malek, K.}, \bibinfo{author}{Gupta, R.}, \bibinfo{author}{Kennedy, D.}, \bibinfo{author}{Swenson, S.C.}, \bibinfo{author}{Wang, L.}, et~al., \bibinfo{year}{2023}.
\newblock \bibinfo{title}{Characterizing uncertainty in community land model version 5 hydrological applications in the united states}.
\newblock \bibinfo{journal}{Scientific Data} \bibinfo{volume}{10}, \bibinfo{pages}{187}.
\bibitem[{Yao et~al.(2023)Yao, Lu, Yang, Xu, Liu, Hu, Yu, Liu, Deng, Tang et~al.}]{ringmo}
\bibinfo{author}{Yao, F.}, \bibinfo{author}{Lu, W.}, \bibinfo{author}{Yang, H.}, \bibinfo{author}{Xu, L.}, \bibinfo{author}{Liu, C.}, \bibinfo{author}{Hu, L.}, \bibinfo{author}{Yu, H.}, \bibinfo{author}{Liu, N.}, \bibinfo{author}{Deng, C.}, \bibinfo{author}{Tang, D.}, et~al., \bibinfo{year}{2023}.
\newblock \bibinfo{title}{Ringmo-sense: Remote sensing foundation model for spatiotemporal prediction via spatiotemporal evolution disentangling}.
\newblock \bibinfo{journal}{IEEE Transactions on Geoscience and Remote Sensing} .
\bibitem[{Yu et~al.(2023)Yu, Sohn, Kim and Shin}]{pvdm}
\bibinfo{author}{Yu, S.}, \bibinfo{author}{Sohn, K.}, \bibinfo{author}{Kim, S.}, \bibinfo{author}{Shin, J.}, \bibinfo{year}{2023}.
\newblock \bibinfo{title}{Video probabilistic diffusion models in projected latent space}, in: \bibinfo{booktitle}{Proceedings of the IEEE/CVF Conference on Computer Vision and Pattern Recognition}, pp. \bibinfo{pages}{18456--18466}.
\bibitem[{Zhang and Zhang(2022)}]{big_data_rs}
\bibinfo{author}{Zhang, L.}, \bibinfo{author}{Zhang, L.}, \bibinfo{year}{2022}.
\newblock \bibinfo{title}{Artificial intelligence for remote sensing data analysis: A review of challenges and opportunities}.
\newblock \bibinfo{journal}{IEEE Geoscience and Remote Sensing Magazine} \bibinfo{volume}{10}, \bibinfo{pages}{270--294}.
\bibitem[{Zhou et~al.(2022)Zhou, Liu, Ding, Fu, Wang, Cai and Shi}]{influence2}
\bibinfo{author}{Zhou, Z.}, \bibinfo{author}{Liu, S.}, \bibinfo{author}{Ding, Y.}, \bibinfo{author}{Fu, Q.}, \bibinfo{author}{Wang, Y.}, \bibinfo{author}{Cai, H.}, \bibinfo{author}{Shi, H.}, \bibinfo{year}{2022}.
\newblock \bibinfo{title}{Assessing the responses of vegetation to meteorological drought and its influencing factors with partial wavelet coherence analysis}.
\newblock \bibinfo{journal}{Journal of Environmental Management} \bibinfo{volume}{311}, \bibinfo{pages}{114879}.
\bibitem[{Zhu et~al.(2022)Zhu, Guo, Deng, Shi, Guan, Zhong, Zhang and Li}]{lulc_change2}
\bibinfo{author}{Zhu, Q.}, \bibinfo{author}{Guo, X.}, \bibinfo{author}{Deng, W.}, \bibinfo{author}{Shi, S.}, \bibinfo{author}{Guan, Q.}, \bibinfo{author}{Zhong, Y.}, \bibinfo{author}{Zhang, L.}, \bibinfo{author}{Li, D.}, \bibinfo{year}{2022}.
\newblock \bibinfo{title}{Land-use/land-cover change detection based on a siamese global learning framework for high spatial resolution remote sensing imagery}.
\newblock \bibinfo{journal}{ISPRS Journal of Photogrammetry and Remote Sensing} \bibinfo{volume}{184}, \bibinfo{pages}{63--78}.

\end{thebibliography}

\end{document}